\documentclass[letterpaper, 10 pt, conference]{ieeeconf}  
\usepackage{graphicx}
\usepackage{siunitx,amssymb,amsmath,bm,amsfonts} 
\usepackage{gensymb}
\graphicspath{{fig/}}
\usepackage{subcaption}
\usepackage{booktabs}
\usepackage{float}
\usepackage[linesnumbered, ruled]{algorithm2e}

\IEEEoverridecommandlockouts                              

\title{\LARGE \bf
Two-Stage Path Following for Mobile Manipulators via Dimensionality-Reduced Graph Search and Numerical Optimization
}

\author{Fuyu Guo$^{1,\dagger}$ , Yuting Mei$^{1,\dagger}$ , Yuyao Zhang$^{1}$ and Qian Tang$^{1}$ 
\thanks{$^{\dagger}$These authors contributed equally to this work.}
\thanks{$^{1}$All Authors are with College of Mechanical and Vehicle Engineering, ChongQing University,
 Chongqing, P.R.C. $\quad$ Contact:  {\tt\small guo$\_$fuyu@icloud.com} $\quad$ {\tt\small tqcqu@cqu.edu.cn}}
}

\begin{document}

\maketitle
\thispagestyle{empty}
\pagestyle{empty}


\begin{abstract}

Efficient path following for mobile manipulators is often hindered by high-dimensional configuration spaces and kinematic constraints.
This paper presents a robust two-stage configuration planning framework that decouples the 8-DoF planning problem into a tractable 2-DoF base optimization under a yaw-fixed base planning assumption.
In the first stage, the proposed approach utilizes IRM to discretize the task-space path into a multi-layer graph, where an initial feasible path is extracted via a Dijkstra-based dynamic programming approach to ensure computational efficiency and global optimality within the discretized graph.
In the second stage, to overcome discrete search quantization, feasible base regions are transformed into convex hulls, enabling subsequent continuous refinement via the L-BFGS algorithm to maximize trajectory smoothness while strictly enforcing reachability constraints.
Simulation results demonstrate the theoretical precision of the proposed method by achieving sub-millimeter kinematic accuracy in simulation, and physical experiments on an omnidirectional mobile manipulator further validate the framework’s robustness and practical applicability.

\end{abstract}

\section{INTRODUCTION}

The integration of robotic arms with mobile bases enhances operational flexibility but introduces a high-dimensional configuration space (C-space) coupling $n$-DoF joint motion with 3-DoF base poses. For complex path-following, traditional inverse kinematics (IK) struggle to balance computational efficiency with optimality over the constructed layered graph, particularly under strict reachability constraints.

To address these challenges, a two-stage planning framework is proposed, leveraging Reachability Maps and Inverse Reachability Maps (RM/IRM). The planning problem is decoupled by first discretizing the task-space trajectory into pose-constrained nodes. For each node, the IRM identifies feasible base configurations to guarantee arm reachability. A multi-layer graph is then constructed from these configuration sets. From this graph, an initial feasible path is extracted via a Dijkstra-based dynamic programming approach, establishing a discrete topological foundation.

In the second stage, these discrete candidates are transformed into continuous geometric feasible regions represented by convex polygons. A numerical optimization based on the L-BFGS algorithm refines the base trajectory to maximize smoothness and ensure configuration continuity. By optimizing within IRM-derived feasible regions, the framework ensures the manipulator operates consistently in high-manipulability zones while strictly satisfying reachability constraints.

\begin{figure}[htbp]
  \centering
  \includegraphics[width=5cm,trim=5cm 4.4cm 1cm 15.5cm,clip]{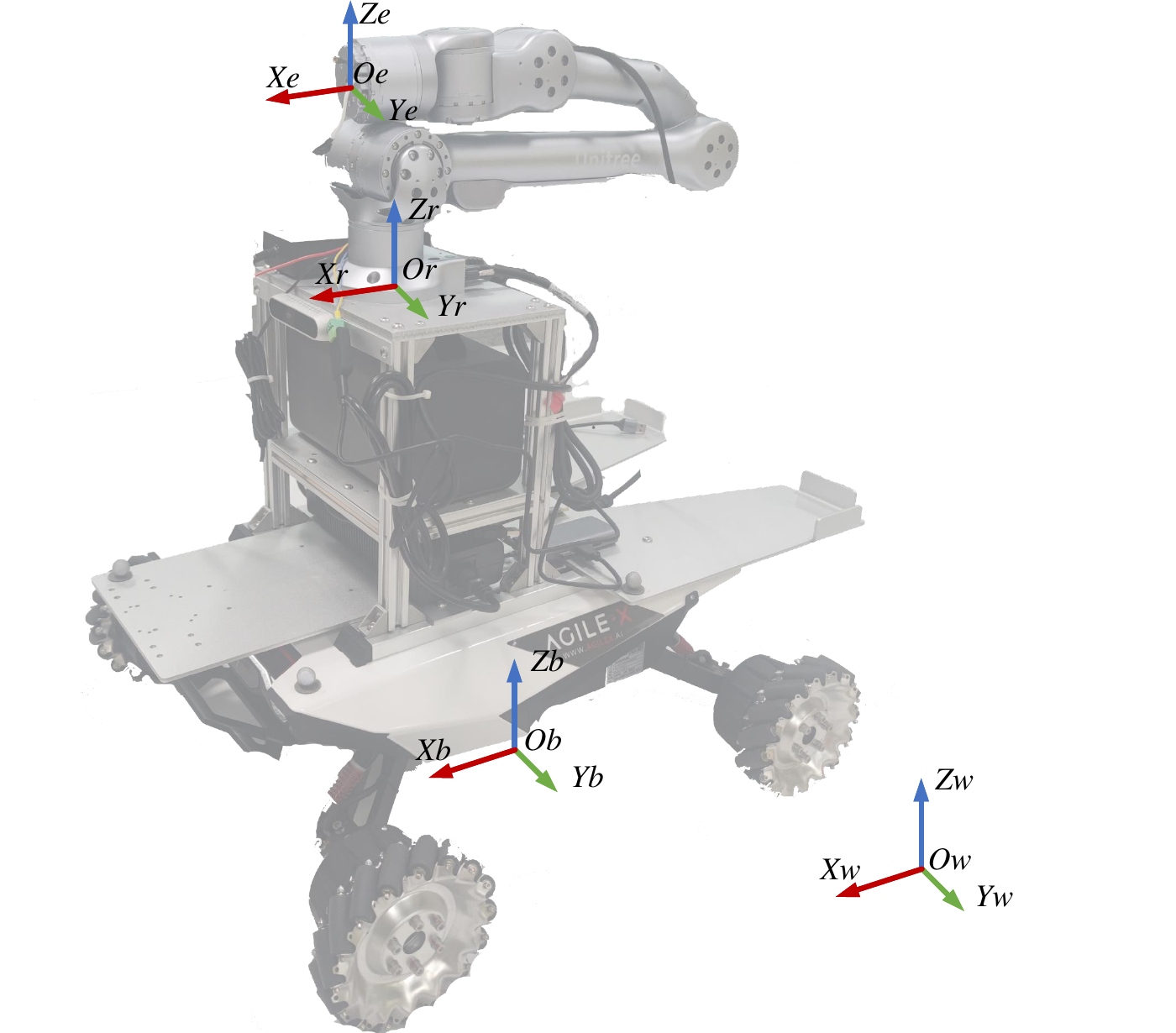}
  \caption{Motion planning for path following of mobile manipulators}
  \label{fig:mobile manipulator}

\end{figure}

The proposed method is validated on an omnidirectional mobile manipulator (Fig. \ref{fig:mobile manipulator}). Experimental results demonstrate that the framework efficiently generates coordinated, smooth trajectories, providing precise guidance for real-world path-following tasks.

\section{RELATED WORK}

Motion planning for path following of mobile manipulators is challenging due to the high-dimensional coupling between base locomotion and arm dexterity \cite{sandakalum2022}. 
Accordingly, planning-based path-following approaches are typically categorized into decoupled and coupled schemes.
More recently, path following has also been achieved using Holistic Reactive Control via pose-based servoing.

\textit{Decoupled and Reachability-based Planning} 
Decoupled approaches simplify computation by planning the base and arm independently.
Early methods leveraged manipulability indices \cite{yoshikawa1985} or separate sampling-based planners (e.g., RRT) for base and arm paths along predefined poses \cite{oriolo2005}.
While computationally fast, these methods often neglect kinematic coupling, leading to infeasible configurations in constrained spaces.
To address this, RM and IRM have been employed to map task-space requirements directly to feasible base regions \cite{zacharias2009, zhang2020}.
Recent works emphasize structured reachability analysis to reduce planning complexity \cite{reister2022, rudorfer2025}, yet many remain limited by the quantization errors of discrete search.
Therefore, a framework capable of translating these discrete feasible spaces into continuous domains for further numerical refinement is highly desirable.

\textit{Coupled and Optimization-based Planning}
Coupled methods treat the MM as a single high-dimensional system. Sampling-based planners (e.g., RRT*) scale well but require intensive post-processing for smoothness. Conversely, optimization-based planners like CHOMP produce high-quality trajectories but are computationally expensive and sensitive to initialization \cite{youakim2018}. Recent hybrid strategies attempt to balance these trade-offs using hierarchical frameworks or reachability-guided sampling \cite{kang2019, ly2024}.

\textit{Holistic Reactive Control:}
Rather than planning the base and arm separately, holistic reactive control treats the mobile manipulator as a unified kinematic system and computes coordinated motions of the base and manipulator in closed loop.
A holistic reactive mobile manipulation framework is proposed in which all base and arm degrees of freedom are jointly optimized within a quadratic program, enabling continuous end-effector servoing while respecting joint limits and preserving maneuverability throughout the motion \cite{haviland2022}.
More recently, reactive base control has been extended to manipulation-on-the-move in dynamic environments by explicitly accounting for static and dynamic obstacles for both the base and arm, thereby improving responsiveness and execution fluidity in sensor-driven tasks \cite{burgess2024}.
While these methods offer strong reactivity and smooth local execution, they primarily focus on instantaneous control objectives and do not provide an explicit global path-structuring mechanism or a discretized feasibility representation for trajectory-wide optimization.

\textit{Our Contribution:} In contrast to coupled optimization methods, which can be computationally demanding, and reactive planners, which typically emphasize local responsiveness over global path structure, the proposed framework first exploits IRM-based discrete search to establish a globally optimal solution within the discretized graph. The resulting discrete feasible sets are then converted into continuous polygonal regions, in which L-BFGS refinement is performed to improve path smoothness while maintaining reachability. Experimental results show that this two-stage strategy provides an effective balance between global feasibility, trajectory smoothness, and practical executability for mobile-manipulator path-following tasks.

\section{Stage One: Discrete Graph Search on Inverse Reachability Maps}
In the first stage of the proposed framework, the task-space trajectory $\mathcal{T}$ is initially discretized, and the Inverse Reachability Map (IRM) is utilized to extract sets of feasible base configurations for each target end-effector pose. Subsequently, these discrete configuration sets are formulated into a multi-layer graph. 
From this constructed graph, a globally optimal base path within the discretized graph is efficiently extracted using a dynamic programming (DP) strategy combined with Dijkstra's algorithm. 
Kinematic reachability is strictly guaranteed by this established discrete topological foundation, which subsequently serves as the initialization for the continuous numerical refinement in the second stage.
\subsection{Reachability Analysis for High-Dimensional Decoupling}
To mitigate the computational burden of high-dimensional configuration planning, a reachability-based mapping between end-effector task-space poses and feasible robot base configurations is precomputed, enabling rapid kinematic feasibility queries and effectively decoupling base planning from repeated IK evaluations and high-dimensional search.
The base orientation is fixed throughout the planning process to simplify IRM construction and reduce the coupling introduced by simultaneous base-arm rotation.

\subsubsection{Reachability Mapping}
The reachability map $\mathcal{R}$ represents the set of all poses reachable by the robot end-effector given a fixed base \cite{zacharias2007}. Reachability maps can be constructed using either forward kinematics sampling or inverse kinematics-based methods \cite{makhal2018}. In this work, the inverse kinematics approach is employed, as it directly provides access to the joint configurations corresponding to each end-effector pose. The reachability map is computed for a Unitree Z1 robotic arm, and the construction procedure consists of the following steps:

\textit{Grid and workspace initialization:}
Discretization resolutions $\delta_p$ and $\delta_r$ are defined for the translational $[x,y,z]$ and rotational $[\alpha,\beta,\gamma]$ components of the end-effector pose, respectively, and the sampling domain is set to a sphere of radius $R$ (equal to the arm's total length) centered at the manipulator base.

\textit{Voxelized workspace generation:}
The 6D pose space is uniformly sampled with $(\delta_p,\delta_r)$ within the spherical workspace, yielding the voxel set
\begin{equation}
\mathcal{V}=\left\{(x,y,z,\alpha,\beta,\gamma)\ \middle|\ x^2+y^2+z^2 \le R^2 \right\}.
\end{equation}

\textit{IK evaluation and operability:}
For each voxel $v\in\mathcal{V}$, IK is solved to obtain a joint configuration $q$; if feasible, an operability score $\mu_k$ is computed and stored in the reachability map as $v_r=(q,\mu_k)$.

\begin{figure}[htbp]
  \centering
  \includegraphics[width=7cm,trim=1cm 12cm 1cm 5cm,clip]{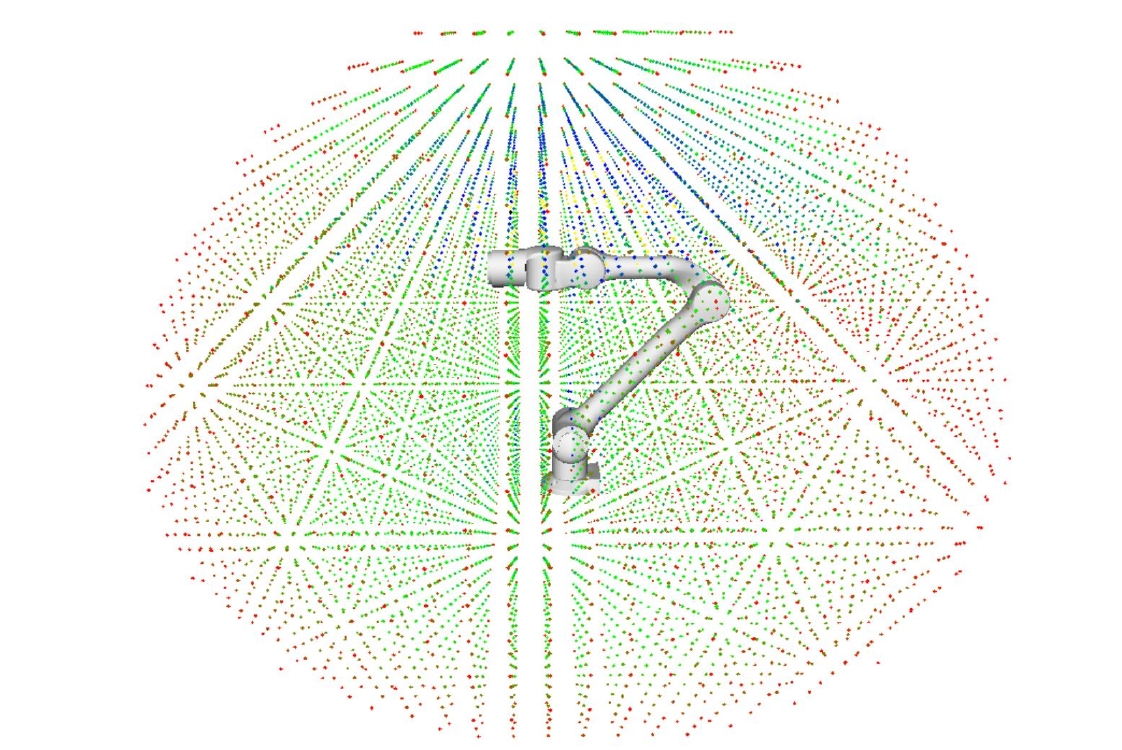}
  \caption{Reachability distribution map of the Z1 robotic arm, where red voxels indicate high pose redundancy and blue voxels signify limited reachability.}
  \label{fig:Reachability distribution map of the Unitree Z1 manipulator}
\end{figure}

\subsubsection{Inverse reachability map projection}
The inverse reachability map $\mathcal{I}$ is constructed by inverting the reachability map $\mathcal{R}$ to enable efficient retrieval of feasible mobile-manipulator configurations for a given end-effector pose $T_e$ \cite{hertle2017}. 
The IRM is generated offline, and its grid is aligned with $\mathcal{R}$ by using the same translational and rotational resolutions $(\delta_p,\delta_r)$. 
A four-level hierarchical index $\mathcal{J}$ is adopted over the pose parameters as
$\mathcal{I}=\mathcal{J}(z,\alpha,\beta,\gamma)$.

Each submap $\mathcal{I}_{k}\in\mathcal{I}$ stores a set of feasible configurations
$\mathcal{Q}_{k}=\{\mathbf{q}_{k1},\mathbf{q}_{k2},\dots,\mathbf{q}_{kn}\}$
associated with the indexed pose bin, where
$
\mathbf{q}_{ki}=\left(x_r, y_r, q_i\right)$,
with $(x_r,y_r)$ denoting the planar base position and $q_i$ the corresponding arm joint configuration.

\textbf{IRM construction:}
Given $\mathcal{R}$, the IRM is built as follows.
\begin{enumerate}
\item A four-level tree index $\mathcal{J}$ over $(z,\alpha,\beta,\gamma)$ is initialized, and an empty container is allocated for each pose bin.
\item For each $v_r\in\mathcal{R}$ with pose $T_e(x,y,z,\alpha,\beta,\gamma)$ and a feasible arm configuration $q_i$, the corresponding base configuration is obtained by inversion as 
$v_i=\left(-x,-y,q_i\right)$.

\item The entry $v_i$ is assigned to the corresponding submap $\mathcal{I}_k$ using the hierarchical index $\mathcal{J}(z, \alpha, \beta, \gamma)$, and the submap is updated as $\mathcal{I}_k \gets \mathcal{I}_k \cup \{v_i\}$.

\item The resulting hierarchical structure is serialized and saved for reuse during online configuration planning.
\end{enumerate}

\subsection{Configuration Planning Algorithm for Path Following}
Configuration planning is formulated as a multi-stage optimal connectivity problem.
Each graph layer is constructed to correspond to the feasible base configurations extracted from the IRM for a specific end-effector pose.
To ensure the extraction of the globally optimal base sequence within the discretized graph, Dijkstra's algorithm is applied in combination with a dynamic programming (DP) strategy, through which the expansion is guided along minimum-cost paths while redundant subproblem computations are avoided.
Base configurations $\mathbf{q}^b_i(x_i,y_i)$ are represented by nodes in the constructed graph, and transition costs are denoted by edges. Both the Euclidean distance and the discrete second derivative (smoothness) between consecutive configurations are minimized by the transition cost:
\begin{equation}
\mathit{length}(i) =
\left\| 
\mathbf{q}^b_{i+1}-\mathbf{q}^b_{i}
\right\|_2
\label{eq:path_length}
\end{equation}
\begin{equation}\mathit{smooth}(i) = 
\left\|
\mathbf{q}^b_{i+1} - 2\mathbf{q}^b_i + \mathbf{q}^b_{i-1} 
\right\|_2
\label{eq:path_smoothness}
\end{equation}

\begin{algorithm}[t]
\caption{Initial configuration planning based on dynamic programming and Dijkstra's algorithm}
\label{alg:Initial Planning}
\KwData{Feasible base configuration set $\mathcal{Q}^b$,initial base configuration $\mathbf{q}^b_{0}$}
\KwResult{Initial base configuration path ${Path}^{b*}$}
$N,m\leftarrow(\mathcal{Q}^b)$\;
$C\gets\infty,P\gets -1$\;
$PQ\gets\{\}$\;
$\mathbf{p}_{base}\gets{\mathbf{q}^b_{0}}$\;
$C[0][i]=0,\{C[0][i],0,0\}\rightarrow PQ$\;
\While{$PQ\ \neq\emptyset\ $}{
	$PQ\rightarrow \mathbf{q}_{head}$\;
	$C_{head},i,j{\gets}\mathbf{q}_{head}$\;

	\If{$C_{head}>\ C[i][j]$}{
		$continue$\;}
	\If{$i<N-1$}{
		\For{$k\in\left[0,\ldots,m_{i+1}-1\right]$}{
			$C_{ij}=length(i)+smooth(i)$\;
			$C_{new}=C\left[i\right]\left[j\right]+C_{ij}$\;
			\If{$C_{new}<C\left[i+1\right]\left[k\right]$}{
				$C\left[i+1\right]\left[k\right]=C_{new}$\;
				$P\left[i+1\right]\left[k\right]=j$\;
				${C_{new},i+1,k}\rightarrow PQ$\;
    				}
  			}
 		}
 	}
\For{$i\in \left[N-1,\ldots,0\right]$}{
	${Path}^{b*} \xleftarrow{+} \mathbf{q}^b_{i}(\mathrm{P})$\;
}

\Return{${Path}^{b*}$}
\end{algorithm}

Based on the defined edge cost formulation and the constructed layered graph, the initial configuration planning algorithm is implemented using dynamic programming combined with Dijkstra search.

Given the feasible base configuration sets $\mathcal{Q}^b$ and an initial configuration $\mathbf{q}^b_{0}$, a cost matrix $C$ is initialized to $\infty$ to record the minimum accumulated cost to each node, and a predecessor matrix $P$ is initialized to $-1$ for path backtracking. A priority queue $PQ$ is used to expand nodes in ascending order of cost. The initial node $\mathbf{q}^b_{0}$ is assigned zero cost and inserted into $PQ$.

During the search, the node with the minimum accumulated cost is extracted from $PQ$, and its transition costs to all feasible configurations in the next stage are evaluated. If a lower cost is found, the corresponding entry in $C$ is updated, the predecessor is recorded in $P$, and the node is inserted into $PQ$. After reaching the final stage, the configuration with the lowest total cost is selected, and the optimal initial base path ${Path}^{b*}$ is recovered by backtracking through $P$.

\begin{figure}[htbp]
  \centering
  \includegraphics[width=6cm,trim=3cm 2cm 3cm 2cm,clip]{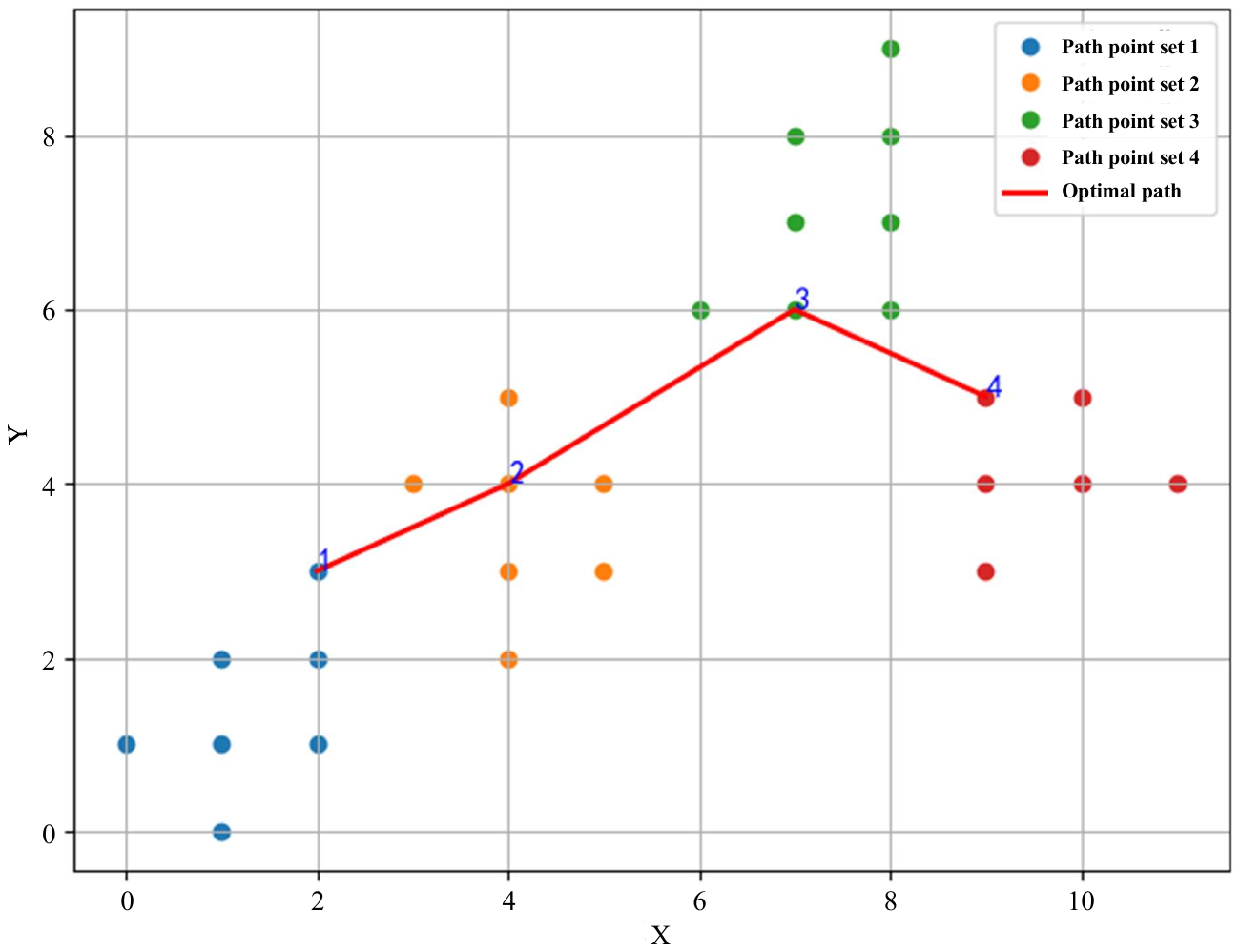}
  \caption{The results of the initial configuration planning method based on dynamic programming and Dijkstra's algorithm}
  \label{fig:Dijkstra-based dynamic programming}
\end{figure}

The details of the proposed graph search strategy are provided in pseudocode form as Algorithm \ref{alg:Initial Planning}.
The results of the graph search demonstrate that a shorter configuration path is achieved while excessive steering angles are avoided, as illustrated in Fig. \ref{fig:Dijkstra-based dynamic programming}. 
Due to the grid spacing constraints of the inverse reachability map, the configuration path generated by this method is characterized by insufficient smoothness.
This issue is subsequently addressed in the second stage through the introduction of continuous refinement via L-BFGS optimization.

\section{Stage Two: Continuous Refinement via L-BFGS Optimization}
In the second stage, to overcome the quantization errors inherent in the discrete graph search, the previously extracted feasible base configurations are transformed into continuous convex regions, enabling subsequent numerical refinement via the L-BFGS algorithm to maximize trajectory smoothness while strongly enforcing reachability constraints through a signed-distance penalty.
\subsection{Geometric Region Representation of Feasible Base Configuration Sets}
\begin{figure}[htbp]
  \centering
  \begin{subfigure}[b]{0.23\textwidth}
    \centering
    \includegraphics[width=\textwidth ,trim=2cm 2cm 3cm 2cm,clip]{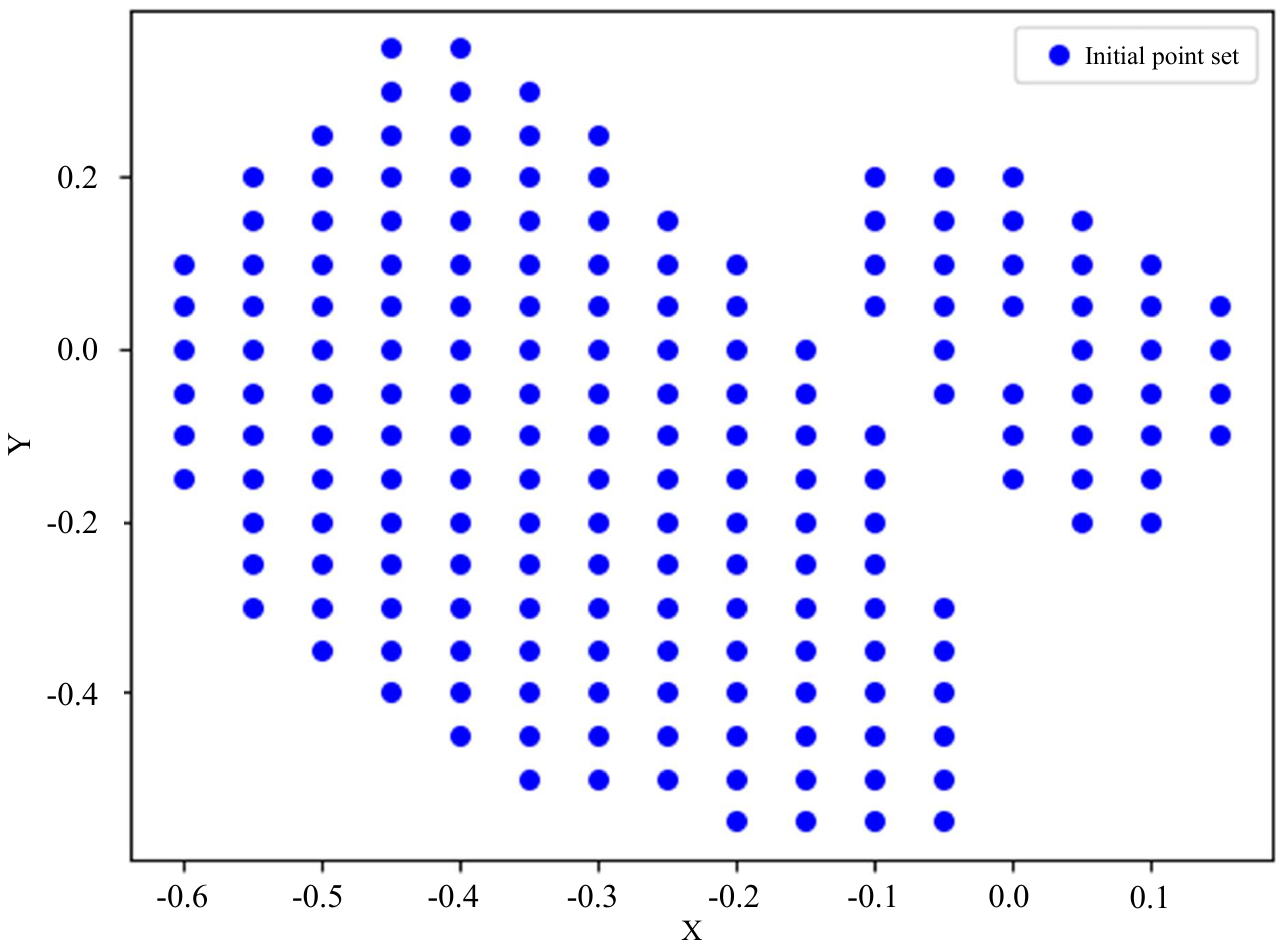}
    \caption{Initial point set}
    \label{fig:InitialPointSet}
  \end{subfigure}
  \hfill
  \begin{subfigure}[b]{0.23\textwidth}
    \centering
    \includegraphics[width=\textwidth ,trim=1cm 2cm 3cm 2cm,clip]{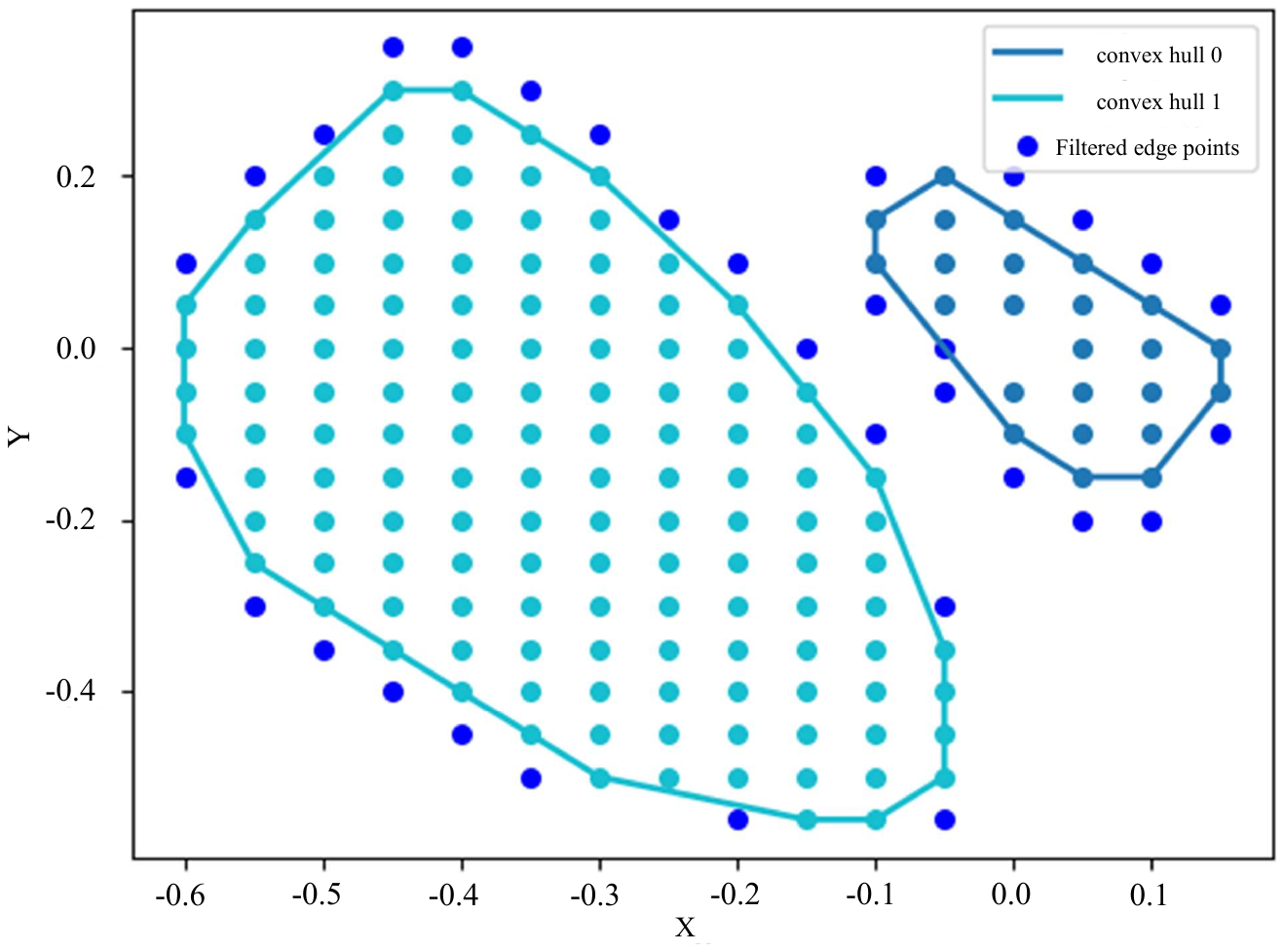}
    \caption{Clustering and partitioning}
    \label{fig:InitialClustering}
  \end{subfigure}

  \vskip 2mm

  \begin{subfigure}[b]{0.23\textwidth}
    \centering
    \includegraphics[width=\textwidth ,trim=0cm 2cm 4cm 2cm,clip]{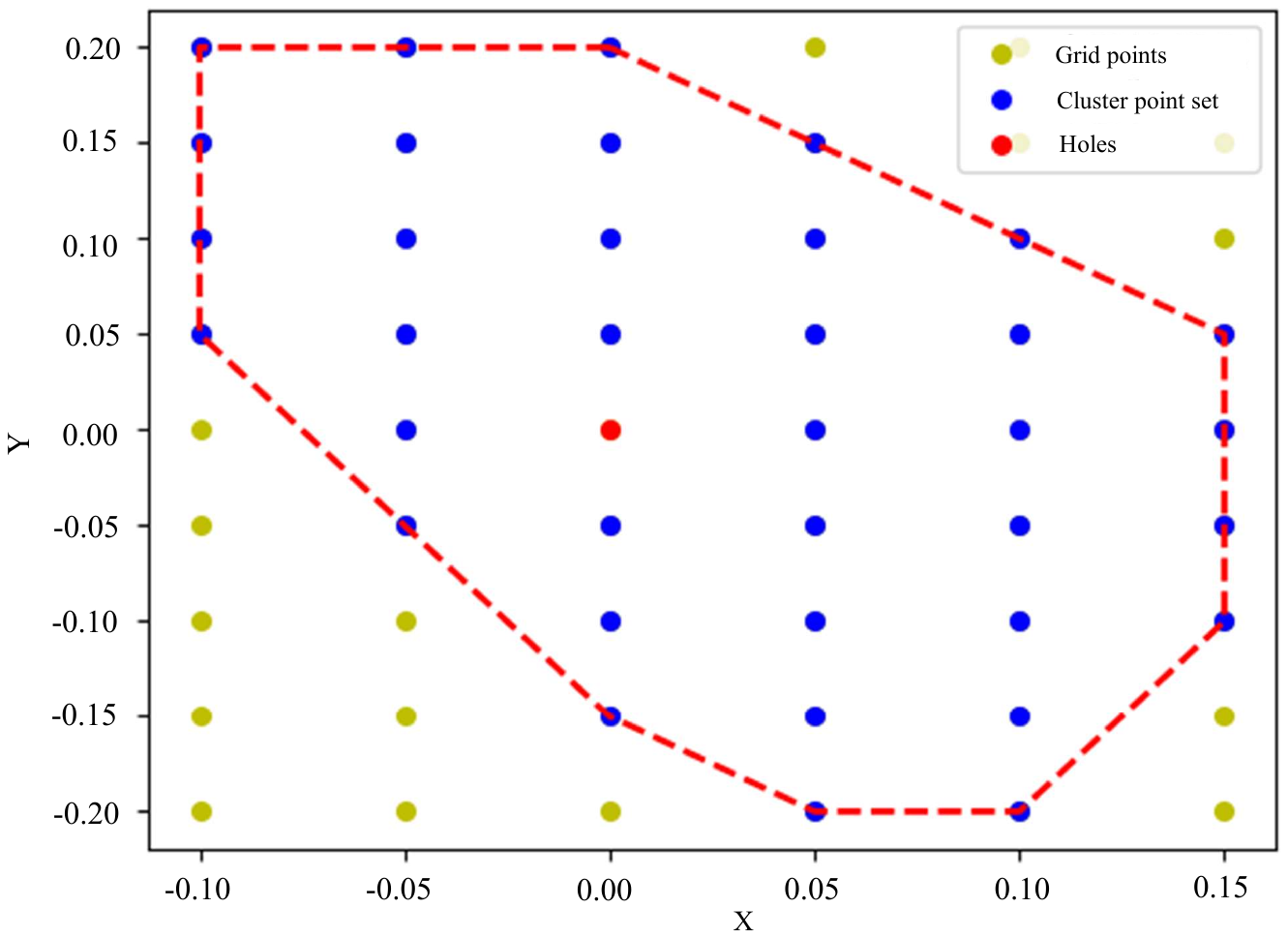}
    \caption{Detection of holes}
    \label{fig:DetectionHoles}
  \end{subfigure}
  \hfill
  \begin{subfigure}[b]{0.23\textwidth}
    \centering
    \includegraphics[width=\textwidth ,trim=0cm 2cm 4cm 2cm,clip]{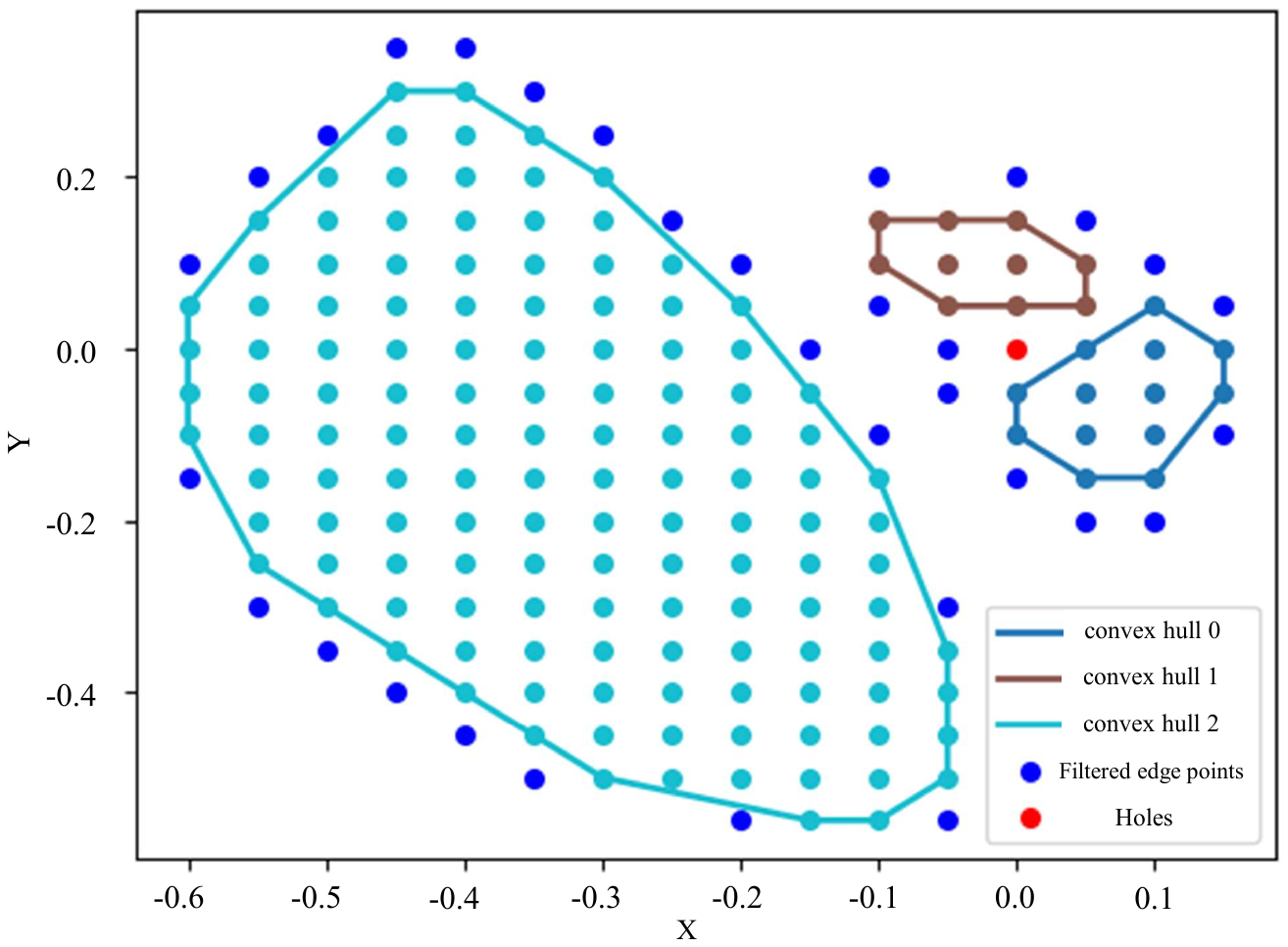}
    \caption{Final convex polygon representation}
    \label{fig:GeometricFeasibleRegions}
  \end{subfigure}

  \caption{Process of converting discrete point sets into geometric feasible regions}
  \label{fig:Converting Geometric Regions}
  
\end{figure}

To facilitate continuous numerical optimization, the discrete feasible base configurations must be transformed into a continuous polygonal representation of feasible sets. This transformation is achieved through a sequential pipeline comprising point set filtering, clustering partitioning, convex hull generation, and recursive hole detection, as detailed in  Algorithm \ref{alg:geom_region_transform} and demonstrated in Fig. \ref{fig:Converting Geometric Regions} .

\begin{algorithm}[t]
\caption{Geometric region representation of discrete configuration points}
\label{alg:geom_region_transform}

\KwData{Initial point set $\mathcal{P}_0$; grid spacing $\delta_g$; minimum cluster size $n_{min}$}
\KwResult{Set of feasible geometric regions $\mathcal{S}$}

$\mathcal{S}\gets\emptyset$ \;
$\mathcal{P}\gets \mathrm{Filter}(\mathcal{P}_0)$ \;
$\mathcal{C}\gets \mathrm{Cluster}(\mathcal{P})$ \;

\For{$\mathcal{C}_i \in \mathcal{C}$}{
    $\mathcal{H}_i \gets \mathrm{ConvexHull}(\mathcal{C}_i)$ \;

    \While{$\mathrm{HasHole}(\mathcal{H}_i,\mathcal{C}_i)$}{
        $\mathcal{C}_i \gets \mathrm{Cluster}(\mathcal{C}_i)$ \;
        $\mathcal{H}_i \gets \mathrm{ConvexHull}(\mathcal{C}_i)$ \;
    }

    \If{$|\mathcal{C}_i| > n_{min}$}{
        $\mathcal{S} \xleftarrow{+} \mathcal{H}_i$ \;
    }
}
\Return{$\mathcal{S}$}
\end{algorithm}

Initially, the feasible base configuration set $\mathcal{Q}^b$ is used as the initial point set $\mathcal{P}_0$. The grid spacing $\delta$ and the minimum cluster size threshold $n_{min}$ are set, and a list of convex polygons without holes $\mathcal{S}\gets\emptyset$ is initialized (Figure \ref{fig:InitialPointSet}). First, point set filtering is performed on $\mathcal{P}_0$ to remove outliers and pseudo-feasible points on the edges:
\begin{align}
\mathcal{P}=\{\mathbf{p}\in \mathcal{P}_0 \mid  |\mathcal{N}(\mathbf{p},\delta_g)| > 2\},
\end{align}
where $\mathcal{N}(\mathbf{p},\delta)$ represents the set of neighboring points within the neighborhood of $\delta$. 
The filtered point set $\mathcal{P}$ is then clustered to generate a sub-cluster set $\mathcal{C} = \{\mathcal{C}_1,\ldots,\mathcal{C}_K\}$, and the convex hull $\mathcal{H}_i=ConvexHull(\mathcal{C}_i)$ of each sub-cluster is calculated (Figure \ref{fig:InitialClustering}). 
Hole detection is performed on each convex hull $\mathcal{H}_i$. If there are no holes, it is added to $\mathcal{S}$; otherwise, the filtering and clustering of sub-clusters are repeated until the number of points $|\mathcal{C}_i|$ is less than the threshold $n_{min}$ (Figure \ref{fig:DetectionHoles}). 
The final list of convex polygons $\mathcal{S}$ is the geometric region representation of the feasible base set at the current path node (Figure \ref{fig:GeometricFeasibleRegions}). 
This method transforms discrete point clouds into continuous and concise convex polygonal geometric regions, which facilitates subsequent cost function calculation.

\subsection{Continuous Configuration Refinement via Numerical Optimization}

To further refine the initial base path, a numerical optimization stage is introduced. The L-BFGS (Limited-memory Broyden-Fletcher-Goldfarb-Shanno) algorithm is employed to optimize the base configuration trajectory while enforcing reachability constraints.

\subsubsection{Cost Function and Gradient Derivation} 
Given a discretized path with $N$ nodes, the configuration optimization problem is formulated as:
\begin{align}
min\sum_{i=0}^{N}{length\left(i\right)+smooth\left(i\right)}\ s.t.\mathbf{q}^b_{i}\in\mathcal{C}_r(i)
\end{align}
where $length(i)$ and $smooth(i)$ denote the length and smoothness costs, respectively, and $\mathcal{C}_r(i)$ represents the convex feasible region associated with node $i$. Specifically, the path length between consecutive nodes is calculated as the Euclidean distance defined in \eqref{eq:path_length}. 
The gradient of the path length with respect to $\mathbf{q}^b_i$ is obtained by accumulating contributions from adjacent segments:
\begin{align}
\frac{\partial \mathit{length}}{\partial \mathbf{q}_{i}^{b}} = \frac{\partial}{\partial \mathbf{q}_{i}^{b}} \Big( \mathit{length}(i) + \mathit{length}(i-1) \Big)
\label{eq:grad_length}
\end{align}
To encourage fluid transitions, a discrete second-order difference term is introduced as shown in \eqref{eq:path_smoothness}.

To apply L-BFGS, the constrained optimization problem is converted into an unconstrained form using a penalty function:
\begin{align}
\min\sum_{i=0}^N{length(i)+smooth(i)+reach(i)}
\end{align}
where $reach(i)$ denotes the feasibility penalty.
Let $d_{\min}(\mathbf{q}^b_i)$ be the minimum distance from 
$\mathbf{q}^b_i$ to the nearest boundary of its associated convex feasible region. 
The penalty term is defined as
\begin{align}
reach(i) = \exp\!\left(\alpha \max\!\bigl(0,\,-d_{\min}(\mathbf{q}_{i}^{b})\bigr)\right),
\label{eq:reach_penalty}
\end{align}
where \(d_{\min}(\mathbf{q}_{i}^{b})\) is the signed distance from \(\mathbf{q}_{i}^{b}\) to the boundary of \(C_r(i)\), defined such that \(d_{\min}(\mathbf{q}_{i}^{b})\ge 0\) inside the feasible region and \(d_{\min}(\mathbf{q}_{i}^{b})<0\) outside. Consequently, the penalty remains constant inside the feasible region and grows exponentially only when the trajectory violates the reachability constraint.
Compared to a quadratic penalty, this exponential formulation provides a superior balance between interior flexibility and boundary enforcement; it maintains a nearly constant cost within the feasible region to allow for unbiased trajectory adjustment, while generating a sufficiently sharp gradient at boundaries to strictly prevent kinematic violations.

The corresponding gradient is
\begin{align}
\frac{\partial\, reach(i)}{\partial \mathbf{q}_{i}^{b}}
&=
\begin{cases}
0, & d_{\min}(\mathbf{q}_{i}^{b}) \ge 0,\\[4pt]
-\alpha \exp\!\left(\alpha(-d_{\min}(\mathbf{q}_{i}^{b}))\right) \\[2pt]
\qquad \qquad \cdot \dfrac{\partial d_{\min}(\mathbf{q}_{i}^{b})}{\partial \mathbf{q}_{i}^{b}},
& d_{\min}(\mathbf{q}_{i}^{b}) < 0.
\end{cases}
\label{eq:reach_grad}
\end{align}

\subsubsection{Configuration Planning Algorithm Based on L-BFGS}

L-BFGS is employed for continuous refinement, where the search direction is computed via a limited-memory quasi-Newton approximation using a small history of curvature pairs $(s_k, y_k)$ and a line search is performed to update the trajectory.

\begin{algorithm}[t]
\caption{L-BFGS-based configuration optimization algorithm}
\label{alg:L-BFGS-optimization}
\KwData{Initial base configuration path ${Path}^{b*}$}
\KwResult{Optimal base configuration path ${Path}^{b}$}
${x}_0\gets{Path}^{b*},H_0^{-1}\gets I,i\gets0 $ \;

\While{$i<iter_{max}$}{
	$\mathbf{d}_i=-H_i^{-1}g\left({x}_i\right)$\;
	$\alpha_i=line\_search(f,g,\mathbf{d}_i)$\;
	${x}_{i+1}={x}_i+\alpha_i \cdot \mathbf{d}_i$\;
	$\bigtriangleup x_i= x_{i+1}-x_i,\bigtriangleup g_i=g(x_{i+1})-g(x_i)$\;
	$H_{i+1}^{-1}=H_i^{-1}+\frac{\bigtriangleup g_i \bigtriangleup g_i^T}{\bigtriangleup g_i^T \bigtriangleup x_i}-\frac{H_i^{-1} \bigtriangleup x_i \bigtriangleup x_i^T H_i^{-1}}{\bigtriangleup x_i^T H_i^{-1} \bigtriangleup x_i}$\;
	\If{$g(x_{i+1})<\delta $}{
	 $Path^b\gets {x}_{i+1}$\;
	\Return{ $Path^b,f\left({x}_{i+1}\right)$\;}
	 }
	$i=i+1$
}

\Return{$False$}
\end{algorithm}

\begin{figure}[htbp]
  \centering
  \includegraphics[width=6cm,trim=0cm 2.2cm 5cm 2cm,clip]{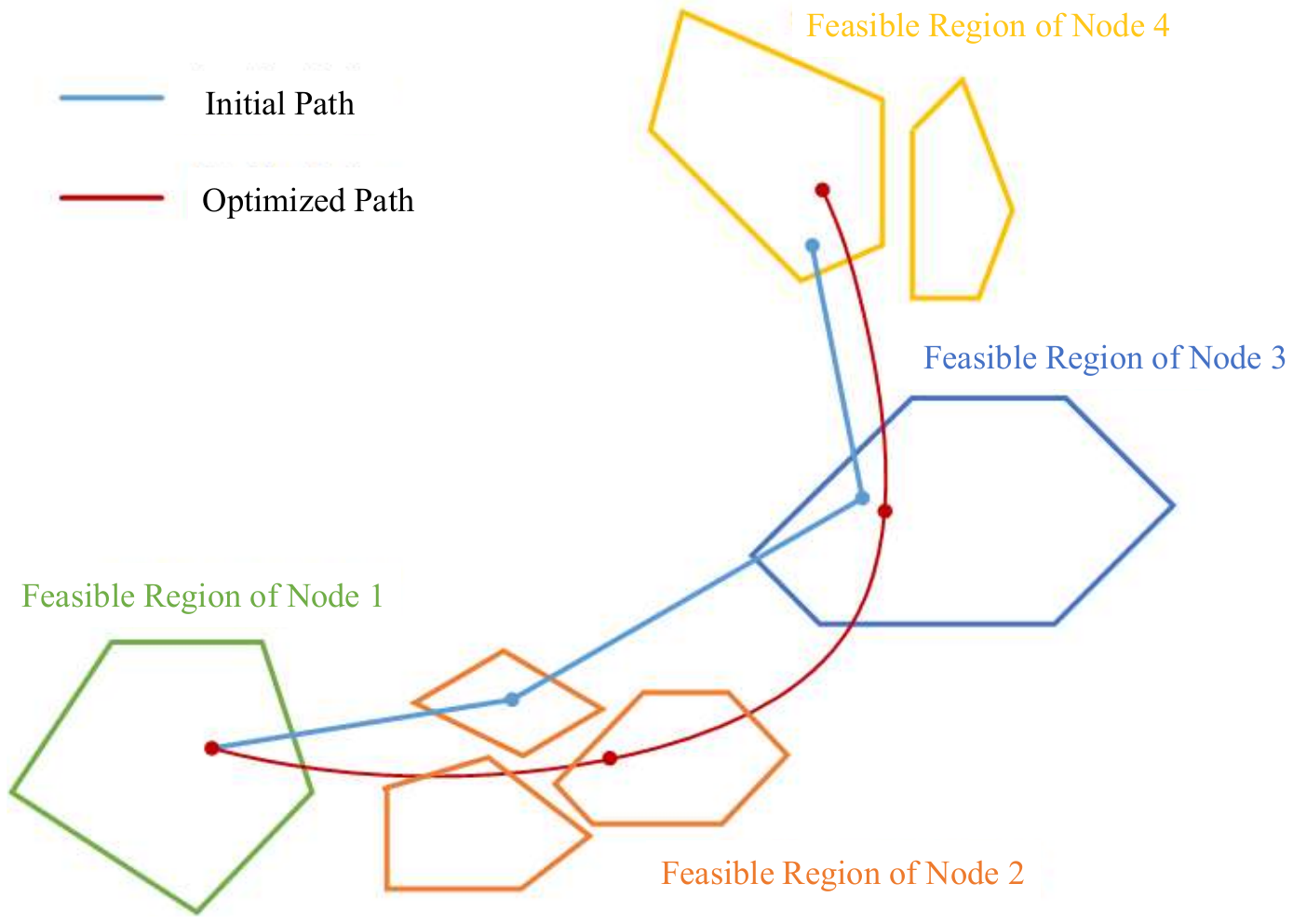}
  \caption{Schematic of secondary configuration optimization}
  \label{fig:L-BFGS-configuration-illustration}

\end{figure}

As shown in Fig.~\ref{fig:L-BFGS-configuration-illustration}, the L-BFGS-based configuration optimization algorithm improves the smoothness of the base configuration path $Path^b$ while ensuring the reachability of the mobile manipulator. Furthermore, by combining the end-effector target trajectory $\mathcal{T}$ and performing kinematic calculations, the complete mobile manipulator configuration path $\mathcal{Q}^r$ can be obtained, providing precise guidance for trajectory tracking tasks.

\section{EXPERIMENTS}
The effectiveness and practical applicability of the proposed two-stage configuration planning method were evaluated through comparative simulations with baseline methods and physical trajectory-tracking experiments on a mobile manipulator platform. 

\subsection{Comparative Evaluation with Holistic Reactive Control}
To evaluate the performance of the proposed two-stage framework, a comparative study was conducted against a standard Holistic Reactive Control (HRC) approach, implemented following the methodology in the Robotics Toolbox for Python \cite{corke2021}. While HRC relies on real-time feedback for local convergence, the proposed method emphasizes global configuration optimality.

The framework was evaluated across five distinct end-effector trajectories ($T_1$--$T_5$) with varying geometric complexities.
The comparison focuses on three key metrics: Base Path Length ($L_b$), Path Smoothness ($S_b$, measured by the integrated squared curvature), and Simulation Kinematic Accuracy ($E_{ee}$, the Euclidean distance between the actual and desired end-effector poses). As detailed in Table~\ref{tab:detailed_comparison}, our approach consistently outperforms HRC in all critical metrics. Notably, the tracking accuracy ($E_{ee}$) of our method remains in the sub-millimeter range ($\approx 10^{-3}$ mm) across all trials, representing a multi-order-of-magnitude improvement over HRC, which exhibits errors up to 14.73 mm. 

\begin{table*}[htbp]
\centering
\caption{Detailed Performance Metrics Across Five Test Trajectories}
\label{tab:detailed_comparison}
\begin{tabular}{lcccccc}
\toprule
Metric & Method & T1 & T2 & T3 & T4 & T5 \\
\midrule
$L_b$ [m] & HRC & $\mathbf{10.84}$ & 4.56 & 8.19 & $\mathbf{6.40}$ & 11.50 \\
          & Two-Stage & 11.23 & $\mathbf{3.93}$ & $\mathbf{7.97}$ & 6.70 & $\mathbf{10.90}$ \\
\addlinespace
$S_b$ [m$^{-1}$] & HRC & $5.65 \times 10^{-2}$ & $9.18 \times 10^{-2}$ & $4.78 \times 10^{-2}$ & $6.03 \times 10^{-2}$ & $5.49 \times 10^{-2}$ \\
          & Two-Stage & $\mathbf{1.12 \times 10^{-2}}$ & $\mathbf{4.32 \times 10^{-4}}$ & $\mathbf{3.26 \times 10^{-3}}$ & $\mathbf{2.48 \times 10^{-2}}$ & $\mathbf{3.03 \times 10^{-3}}$\\
\addlinespace
$E_{ee}$ [mm] & HRC & 0.643 & 6.193 & 7.785 & 6.780 & 14.734 \\
          & Two-Stage & $\mathbf{0.0014}$ & $\mathbf{0.0017}$ & $\mathbf{0.0012}$ & $\mathbf{0.0014}$ & $\mathbf{0.0019}$ \\
\bottomrule

\end{tabular}
\end{table*}

\begin{figure}[htbp]
  \centering
  \begin{subfigure}[b]{0.23\textwidth}
    \centering
    \includegraphics[width=\textwidth ,trim=4.5cm 2cm 4.5cm 2cm,clip]{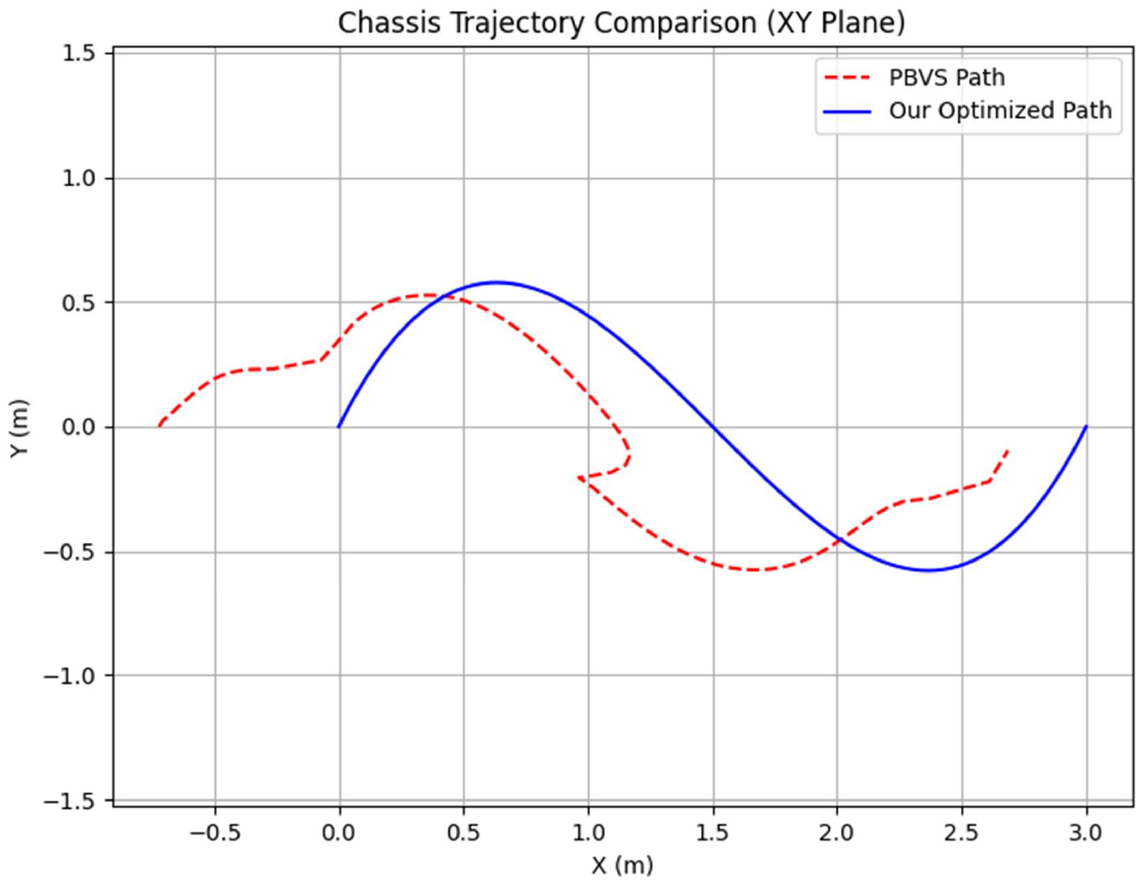}
    \caption{Base Trajectory Comparison(XY Plane)}
    \label{fig:base_s_curve}
  \end{subfigure}
  \hfill
  \begin{subfigure}[b]{0.23\textwidth}
    \centering
    \includegraphics[width=\textwidth ,trim=5cm 2cm 5cm 2cm,clip]{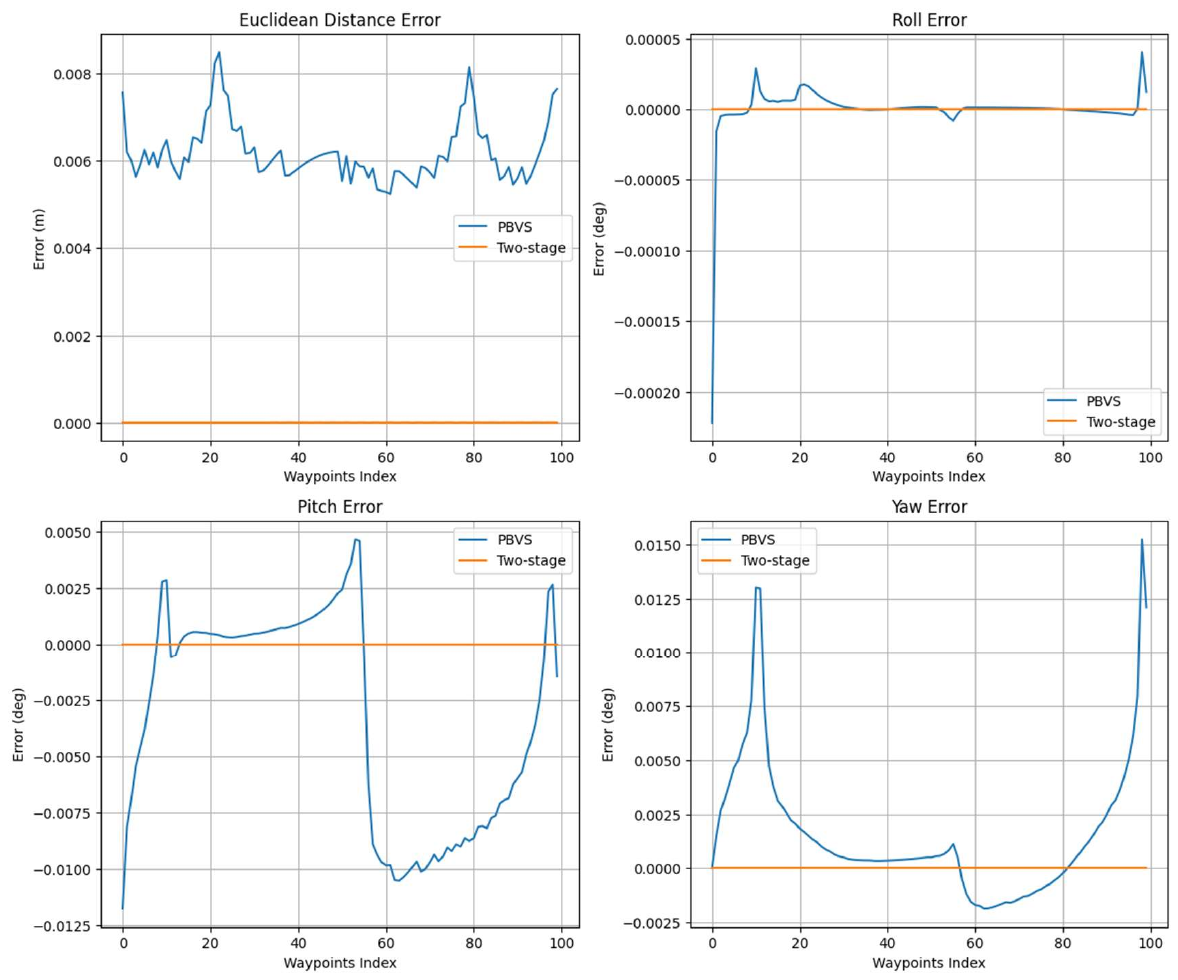}
    \caption{End-effector Trajectory Comparison}
    \label{fig:error_s_curve}
  \end{subfigure}

  \caption{Comparison of trajectory tracking performance between HRC and the proposed method.}
  \label{fig:s_curve}

\end{figure}

The superiority in path smoothness ($S_b$) is equally significant. Our L-BFGS-based refinement achieves smoothness values that are consistently one to two orders of magnitude lower than the baseline. This is further illustrated in Fig.~\ref{fig:s_curve}, which depicts a representative S-shaped trajectory tracking trial. In Fig.~\ref{fig:base_s_curve}, the HRC-controlled base exhibits characteristic "jerky" or oscillatory behavior as it struggles to maintain the arm's reachability. In contrast, our method generates a proactive and fluid base trajectory within the IRM-defined feasible regions. By ensuring the manipulator consistently operates within high-manipulability configurations, the proposed method provides a more stable and efficient solution for high-precision mobile manipulation tasks.

\subsection{Comparative Analysis with CHOMP}

To further validate the trajectory fidelity of the proposed framework, a comparative study was conducted against a CHOMP-based optimization approach \cite{zucker2013chomp}. While CHOMP is effective for generating collision-free movements, it treats path adherence as a soft-constrained cost term, often leading to observable drift in high-precision tasks.

As summarized in Table~\ref{tab:rmse_comparison}, the proposed method demonstrates a decisive improvement in tracking precision across all five test trajectories ($T_1$--$T_5$). The Root Mean Square Error (RMSE) of the end-effector position for our framework remains consistently within the sub-micrometer to sub-millimeter range ($\approx 10^{-4}$ to $10^{-3}$ mm). In contrast, the CHOMP-based method exhibits significantly higher RMSE values, particularly in $T_3$ where the error reaches 8.6882 mm. This discrepancy arises because CHOMP tends to "cut corners" to minimize the smoothness functional $f_{smooth}$, whereas our framework utilizes an exponential penalty function $e^{\alpha \cdot d_{\min}}$ to generate a sharp gradient at the boundaries of the feasible configuration space. This ensures that the optimizer strictly enforces kinematic reachability and path consistency, effectively neutralizing the tracking drift inherent in standard gradient-descent planners.

\begin{table}[htbp]
\centering
\caption{End-Effector Position RMSE Comparison (mm)}
\label{tab:rmse_comparison}
\begin{tabular}{lccccc}
\toprule
Method & T1 & T2 & T3 & T4 & T5 \\
\midrule
CHOMP & 0.9017 & 1.3804 & 8.6882 & 1.4459 & 0.6243 \\
Two-Stage & $\mathbf{0.0014}$ & $\mathbf{0.0012}$ & $\mathbf{0.0001}$ & $\mathbf{0.0016}$ & $\mathbf{0.0008}$ \\
\bottomrule
\end{tabular}
\end{table}

\subsection{Calibration of the Mobile Manipulator}
All hardware experiments were carried out on a mobile manipulator consisting of a Unitree Z1 robotic arm and an Agilex Scout Mini Mecanum-wheeled base. The system calibration and trajectory tracking evaluation were conducted with the aid of a NOKOV Mars optical 3D motion capture system.
Accurate kinematic modeling of the mobile manipulator requires precise estimation of key structural parameters. A motion capture system was employed to calibrate both the base center and the installation position of the robotic arm. High-precision pose measurements from the motion capture system were used to determine the parameters necessary for kinematic modeling and trajectory planning.

\subsubsection{Calibration of the Base Center}

Since a reflective marker cannot be mounted exactly at the geometric center of the base, the platform was rotated in place while recording the trajectories of multiple markers attached to the base. Each marker trajectory was fitted to a circle, and the average of the fitted circle centers was taken as the kinematic center of the mobile base, as illustrated in Fig.~\ref{fig:BaseCenter}. The estimated center coordinates are summarized in Table~\ref{tab:base_center}.

\begin{table}[htbp]
\centering
\caption{Center Coordinates of the Fitted Circular Trajectories}
\label{tab:base_center}
\begin{tabular}{cccccc}
\toprule
Marker ID & 1 & 2 & 3 & 4 & Mean \\
\midrule
$x$ (m) & 0.62175 & 0.62460 & 0.62258 & 0.62225 & 0.62280 \\
$y$ (m) & 0.04963 & 0.03952 & 0.04194 & 0.04059 & 0.04292 \\
\bottomrule

\end{tabular}
\end{table}

\subsubsection{Calibration of the Robotic Arm Installation Position}
\begin{figure}[htbp]
  \centering
  \begin{subfigure}{0.47\linewidth}
    \centering
    \includegraphics[width=\linewidth ,trim=4cm 2cm 5cm 1cm,clip]{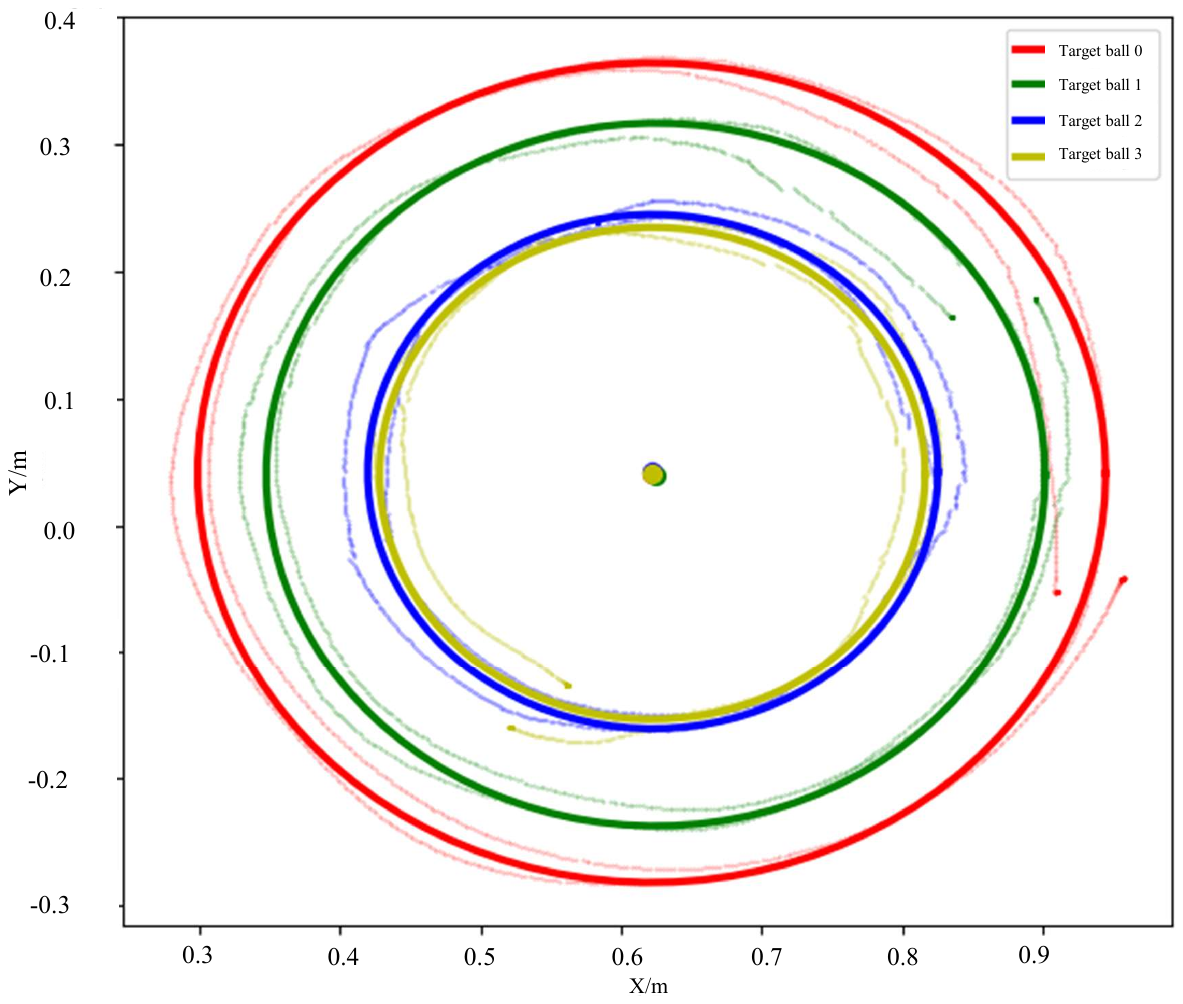}
    \caption{Trajectories of target balls and center positions of the rotating base}
    \label{fig:BaseCenter}
  \end{subfigure}
  \hfill
  \begin{subfigure}{0.47\linewidth}
    \centering
    \includegraphics[width=\linewidth ,trim=4cm 1cm 7cm 1cm,clip]{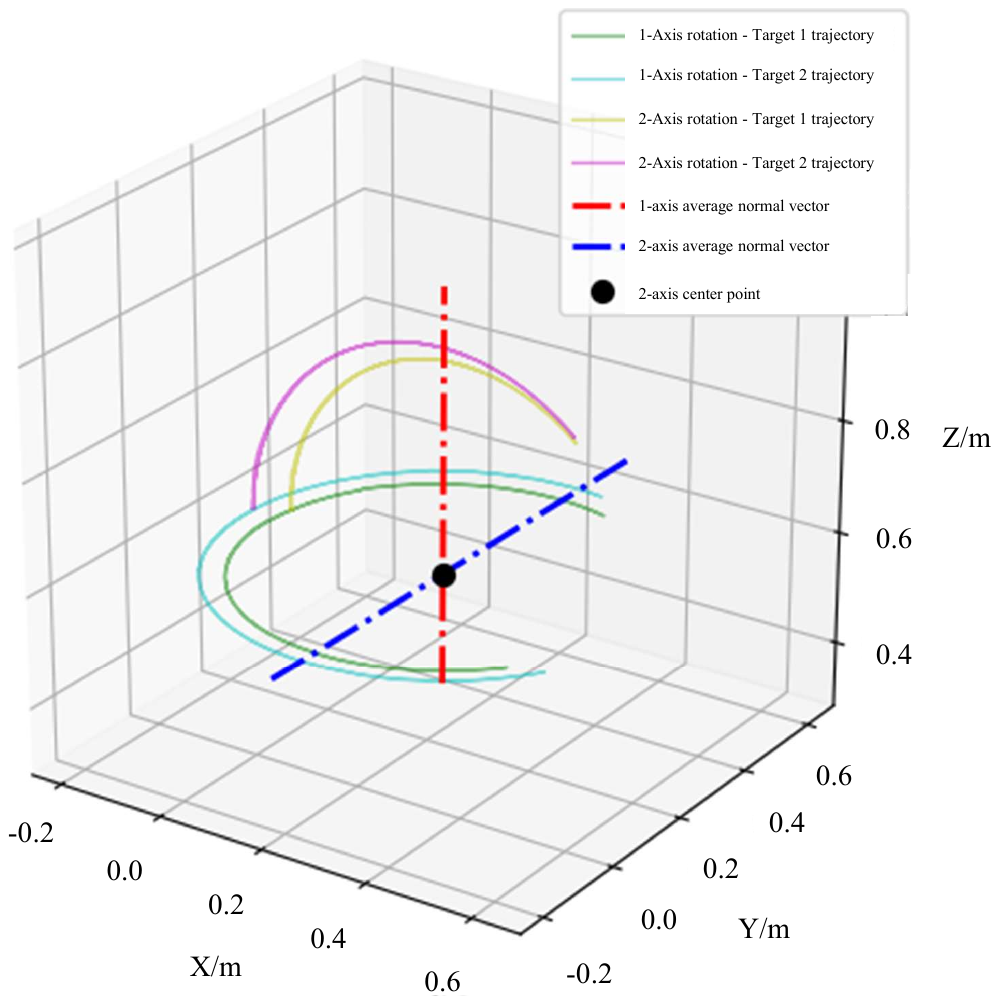}
    \caption{Calibration of the axis position for the robotic arm's second axis}
    \label{fig:ArmBaseAxis2}
  \end{subfigure}
  \caption{Kinematic calibration of the mobile manipulator system.}
\end{figure}

To determine the mounting position of the robotic arm relative to the base center, reflective markers were attached to the end-effector and intermediate links. The first and second joints were rotated independently, and the rotational centers of the tracked markers were computed to identify the joint axes. The intersection of these axes yields the position of the second joint axis, as illustrated in Fig.~\ref{fig:ArmBaseAxis2}. The calibrated position of the second joint axis under the current base configuration is
(0.24644~\text{m}, ~0.17876~\text{m}, ~0.60475~\text{m}).

Assuming only a translational transformation exists between the mobile base coordinate frame and the robotic arm base frame, the current base pose can be computed using the calibrated base center and the arm model parameters. Consequently, the homogeneous transformation matrix from the base frame to the arm base frame, denoted as $T_{br}$, is obtained as:

\begin{align}
T_{br}= \begin{bmatrix} 1 & 0 & 0 & 0.08764 \\ 0 & 1 & 0 & 0.00214 \\ 0 & 0 & 1 & 0.50125 \\ 0 & 0 & 0 & 1 \end{bmatrix} 
\end{align}

\subsection{Trajectory Tracking Accuracy Evaluation}
Experiments were performed on three target paths of different shapes, namely a lemniscate, a capsule-shaped curve, and a polygonal path, to validate the feasibility of the proposed configuration planning approach for path following.

\begin{figure}[htbp]

  \centering
  \begin{subfigure}[b]{0.235\textwidth}
    \centering
    \includegraphics[width=\textwidth, trim=1cm 2cm 10cm 2cm, clip]{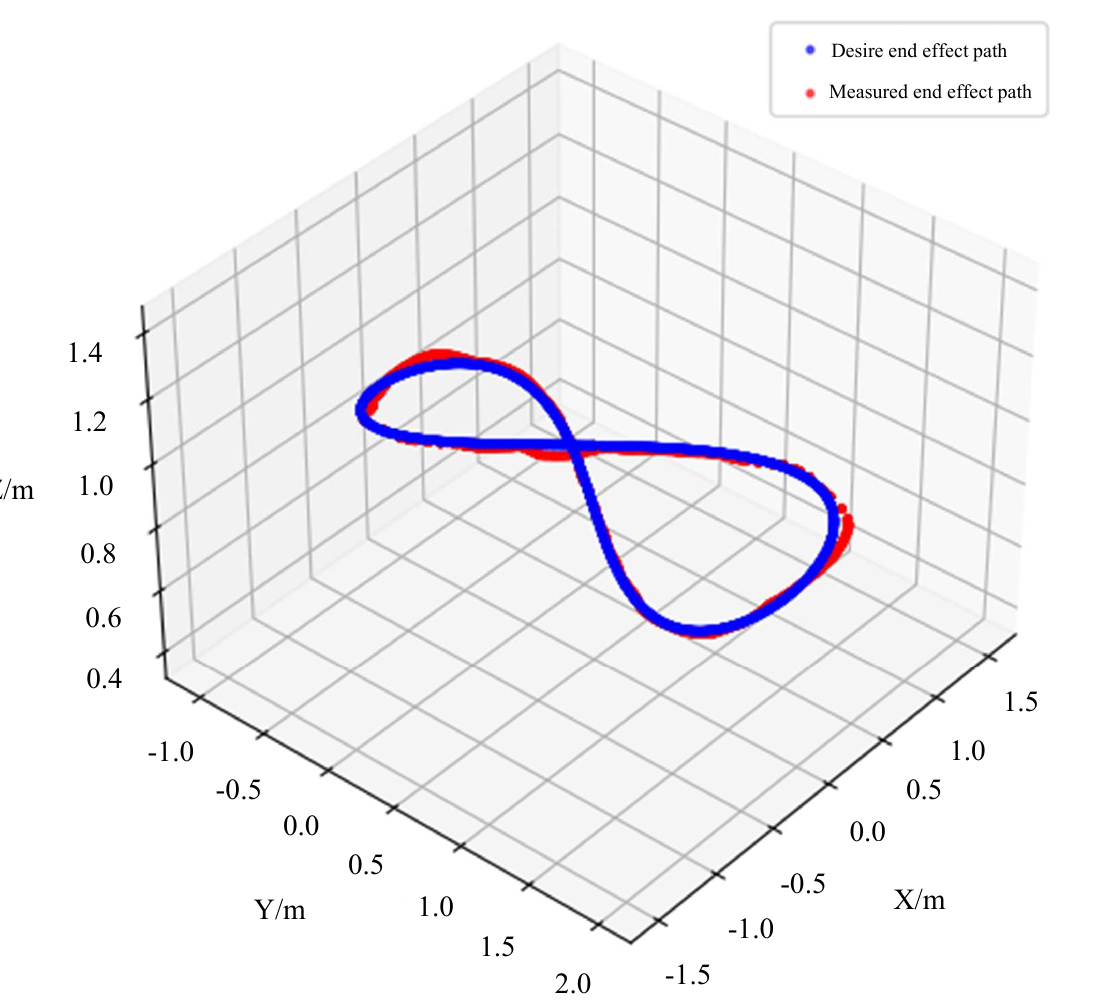}
    \caption{Trajectory}
  \end{subfigure}
  \hfill
  \begin{subfigure}[b]{0.235\textwidth}
    \centering
    \includegraphics[width=\textwidth, trim=0cm 0cm 2cm 1cm, clip]{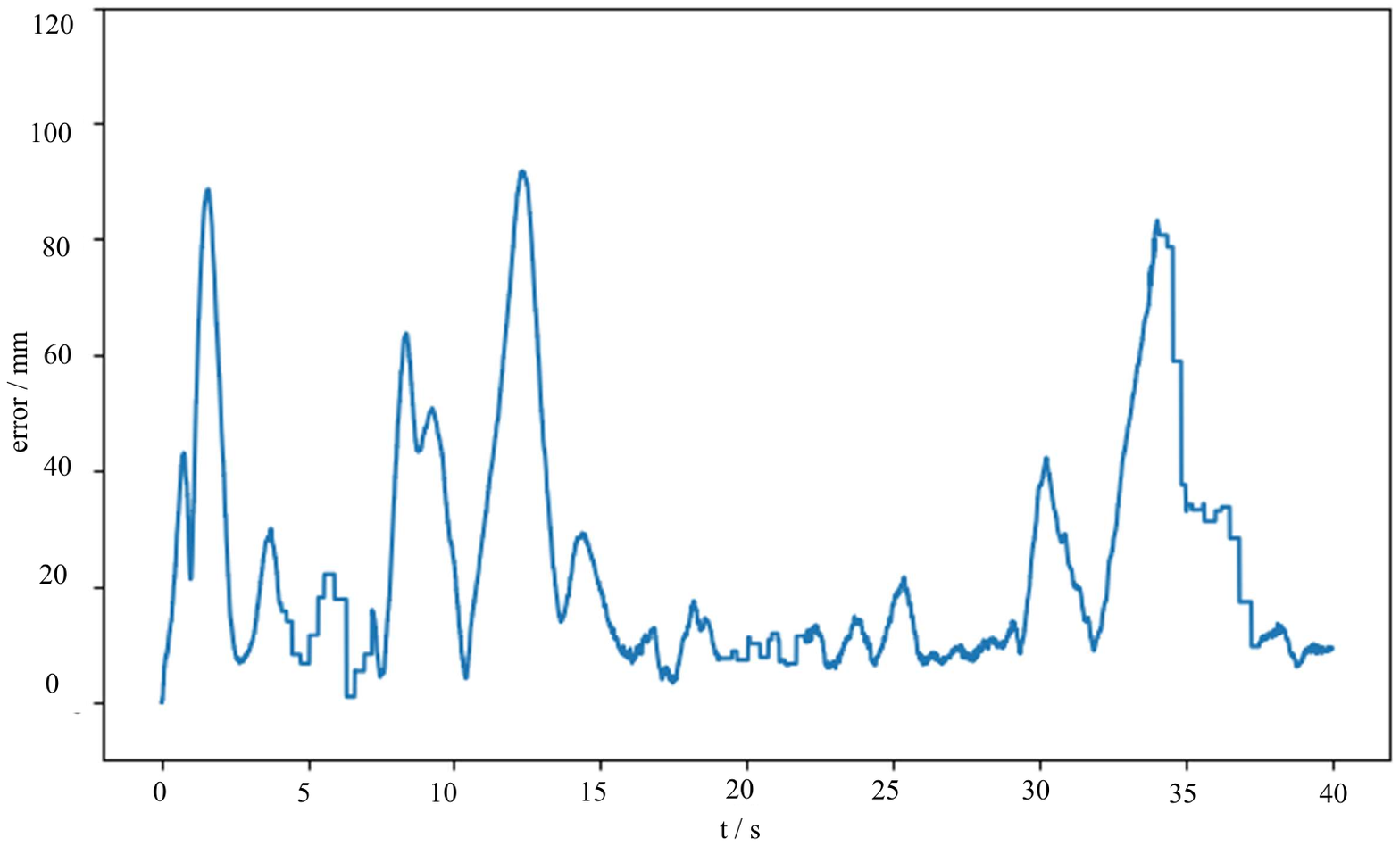}
    \caption{Tracking error}
  \end{subfigure}

  \caption{Tracking performance for the lemniscate path.}
  \label{fig:lemniscate_results}

\end{figure}

\begin{figure}[htbp]

  \centering
  \begin{subfigure}[b]{0.235\textwidth}
    \centering
    \includegraphics[width=\textwidth, trim=3cm 1.5cm 8cm 2cm, clip]{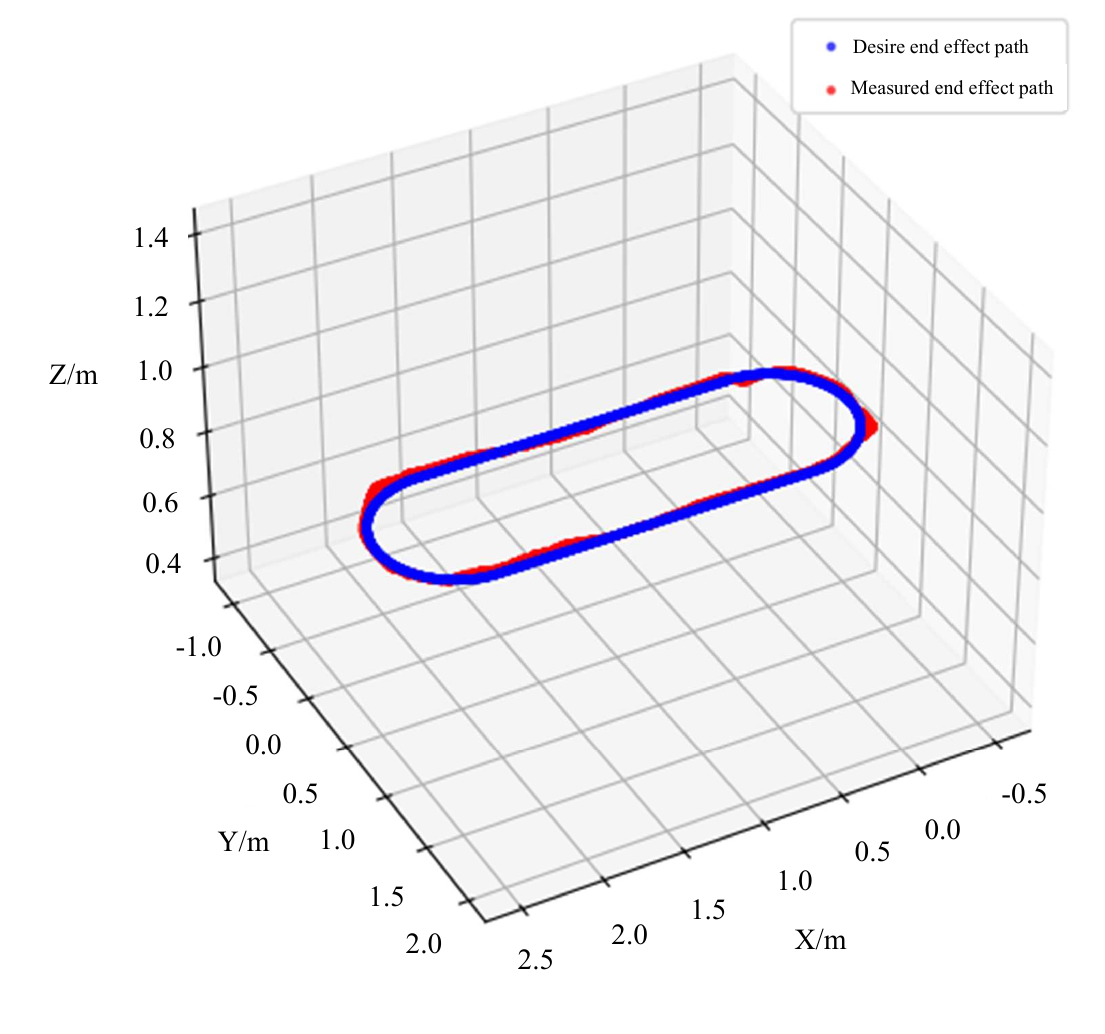}
    \caption{Trajectory}
  \end{subfigure}
  \hfill
  \begin{subfigure}[b]{0.235\textwidth}
    \centering
    \includegraphics[width=\textwidth, trim=0cm 0cm 0cm 2cm, clip]{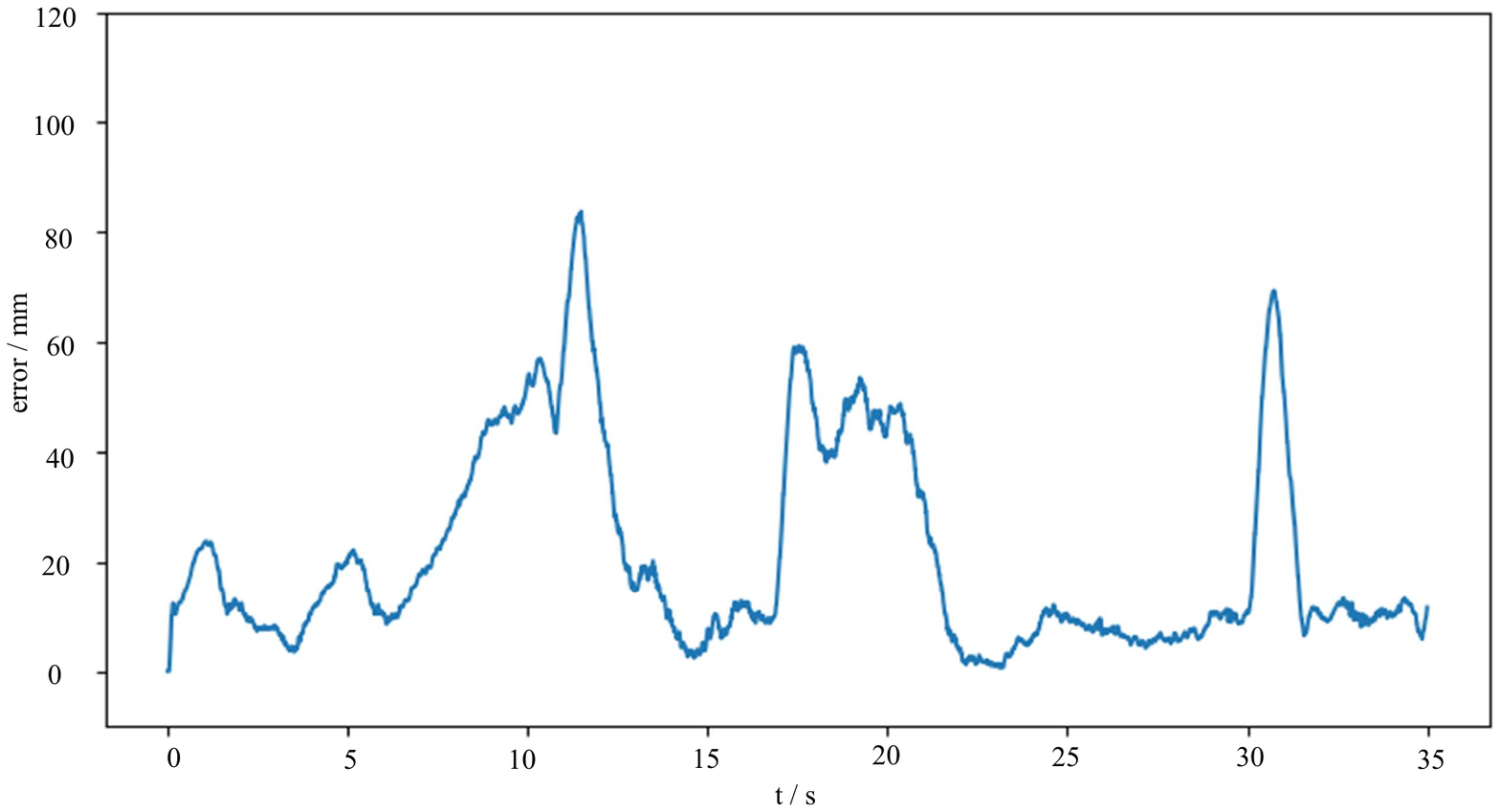}
    \caption{Tracking error}
  \end{subfigure}

  \caption{Tracking performance for the capsule path.}
  \label{fig:capsule_results}

\end{figure}

\begin{figure}[htbp]
  \centering
  \begin{subfigure}[b]{0.235\textwidth}
    \centering
    \includegraphics[width=\textwidth, trim=3cm 1.5cm 10cm 2cm, clip]{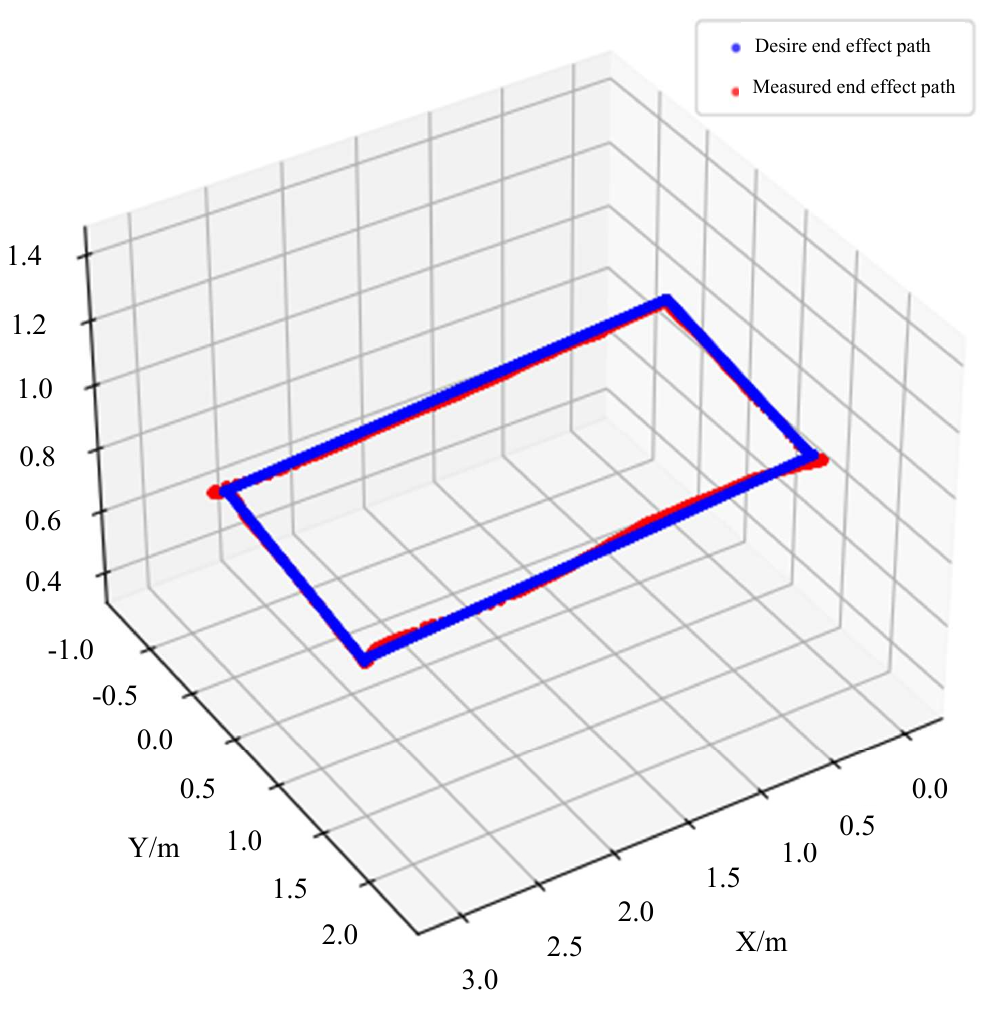}
    \caption{Trajectory}
  \end{subfigure}
  \hfill
  \begin{subfigure}[b]{0.235\textwidth}
    \centering
    \includegraphics[width=\textwidth, trim=0cm 0cm 0cm 2cm, clip]{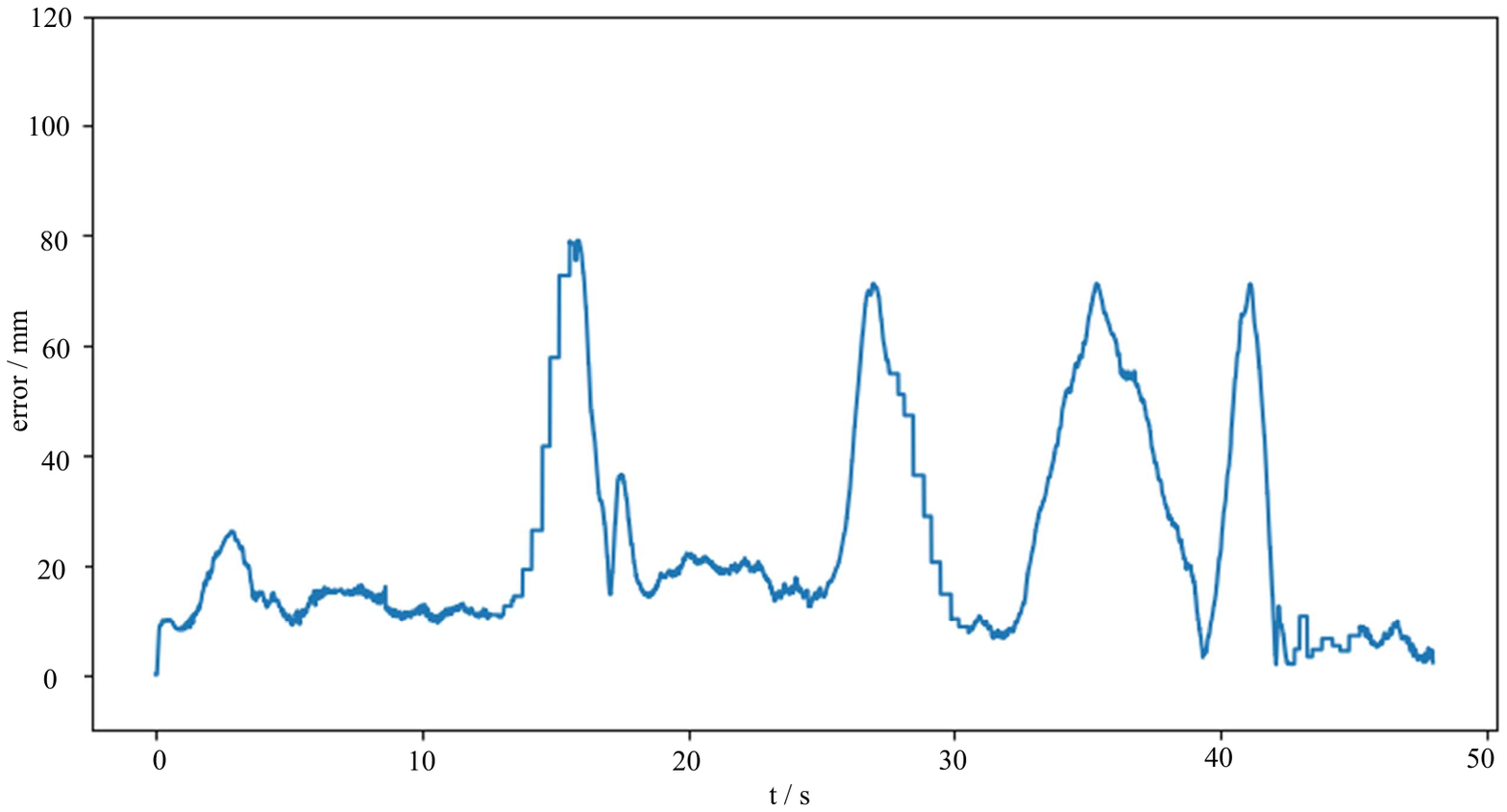}
    \caption{Tracking error}
  \end{subfigure}

  \caption{Tracking performance for the polygonal path.}
  \label{fig:polygonal_results}

\end{figure}

\begin{table*}[t]
\centering
\caption{Tracking Accuracy for Different Trajectories}
\label{tab:tracking_accuracy}
\begin{tabular}{lcccc}
\toprule
Trajectory & Dimensions (m) & Max Error (mm) & Mean Error (mm) & Error Location \\
\midrule
Lemniscate & (1.2, 3.0, 0.2) & 91.82 & 23.26 & High curvature \\
Capsule-shaped & (2.5, 1.0, 0) & 83.72 & 21.17 & Curve transitions \\
Polygonal & (3.0, 1.5, 0.2) & 80.65 & 20.03 & Corners \\
\bottomrule
\end{tabular}
\end{table*}

An omnidirectional Mecanum-wheeled chassis was selected for its kinematic simplicity and ability to follow arbitrary trajectories, facilitating rapid algorithmic verification. The actual end-effector trajectories were recorded using a motion capture system and compared with the desired trajectories. However, the inherent limitations of Mecanum wheels—such as complex slip dynamics and limited load capacity—contribute significantly to the tracking deviations observed in real-world deployments.
As summarized in Table~\ref{tab:tracking_accuracy}, the mean tracking error is approximately 21--23 mm. The maximum deviations primarily occur at high-curvature segments or corner transitions, where wheel-ground interaction and non-linear slip effects are most pronounced. While these hardware-specific factors limit absolute precision, the experiments successfully demonstrate the feasibility of the two-stage configuration planning. To mitigate these dynamic errors in industrial applications, future work will involve migrating the algorithm to a four-wheel independent steering (4WIS) platform with superior traction and load stability.
To the best of our knowledge, the AgileX Ranger Mini 4WIS platform, which is widely used in both academia and industry, does not provide a unified general $\mathrm{SE}(2)$ control interface that simultaneously combines in-place rotation and translational motion. Accordingly, the subsequent experiments on the 4WIS platform will be conducted under a crab-motion model, in which the base orientation is kept fixed while planar translation is executed.
This choice is consistent with the non-rotating assumption adopted in the proposed planning framework.

\begin{figure}[htbp]
  \centering
  \begin{subfigure}[b]{0.23\textwidth}
    \centering
    \includegraphics[width=\textwidth ,trim=5cm 2.3cm 6cm 2cm,clip]{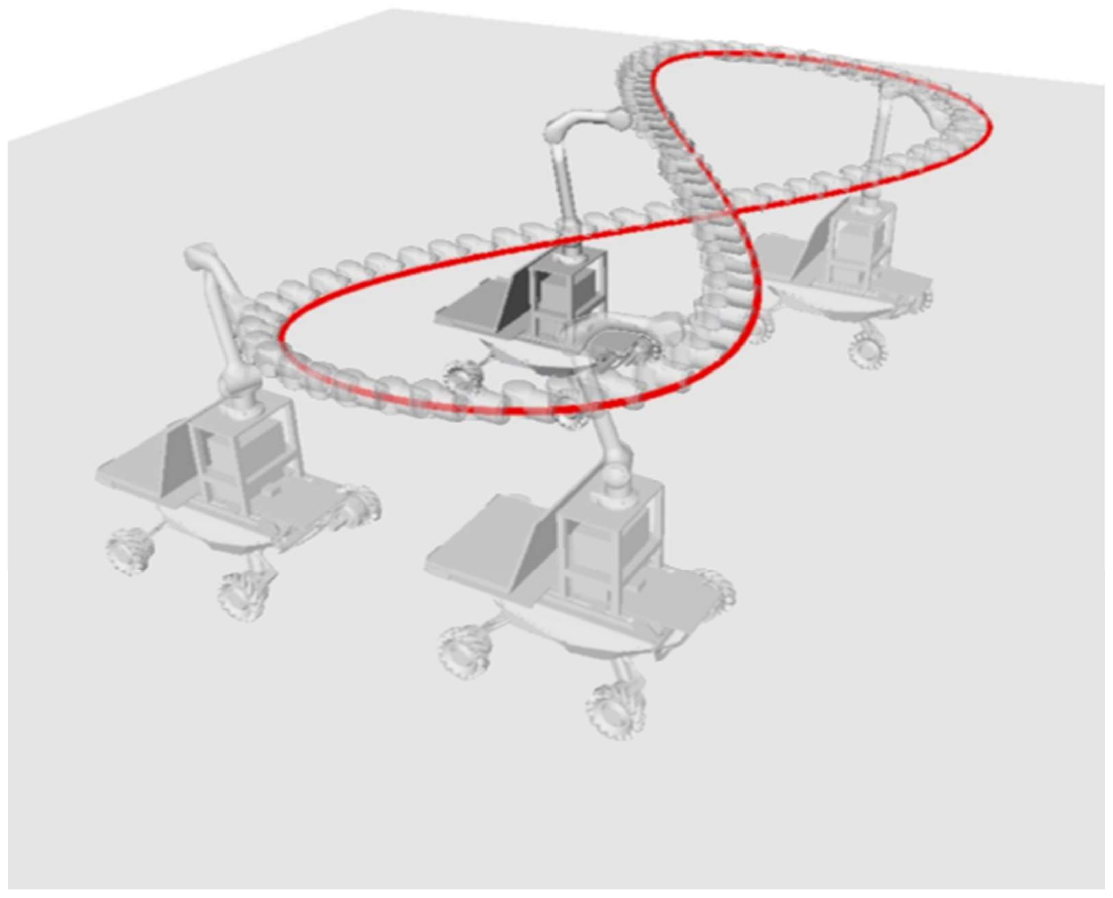}
    \caption{Simulation configuration path}
    \label{fig:SimulationConfigurationPath}
  \end{subfigure}
  \hfill
  \begin{subfigure}[b]{0.23\textwidth}
    \centering
    \includegraphics[width=\textwidth ,trim=5cm 2cm 5cm 2cm,clip]{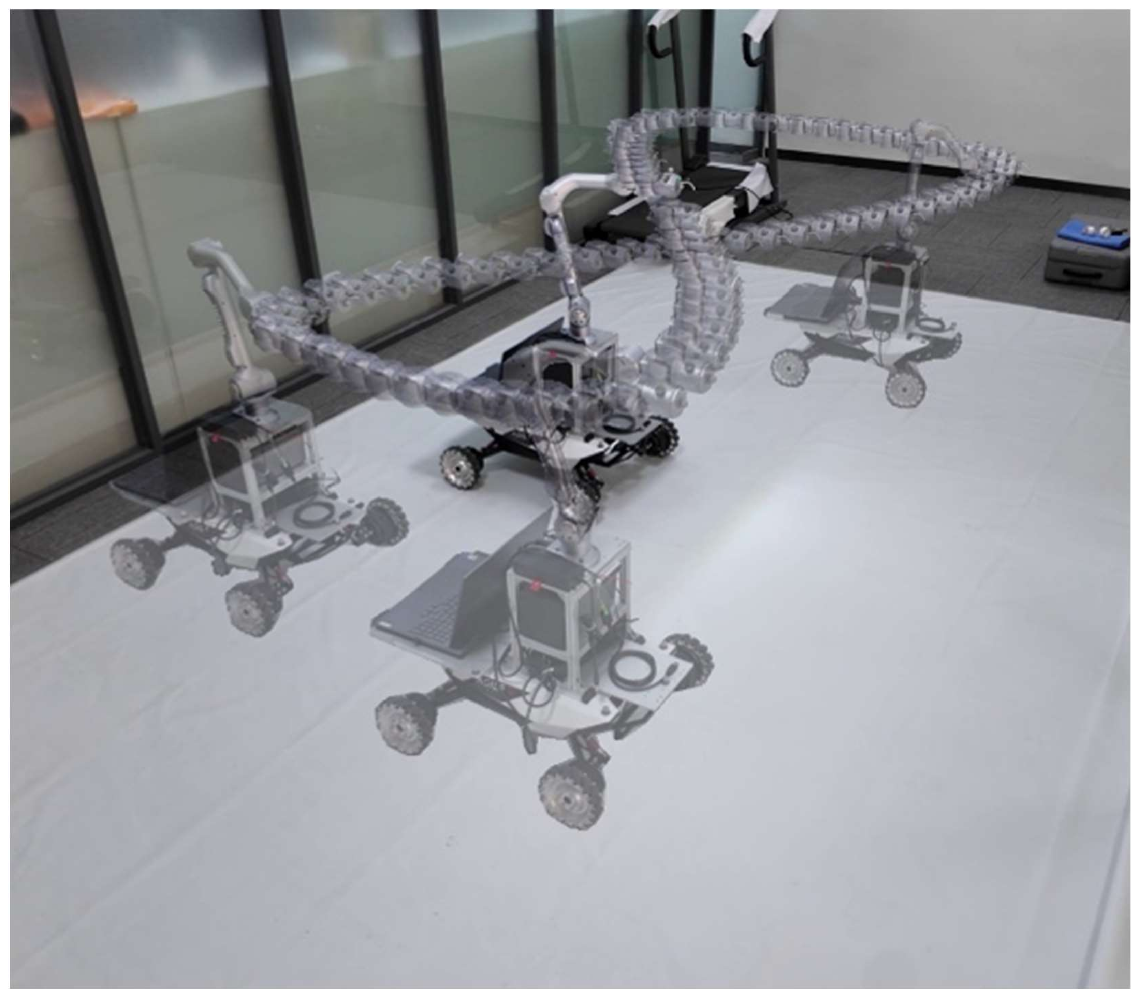}
    \caption{Experimental configuration path}
    \label{fig:ExperimentalConfigurationPath}
  \end{subfigure}

  \caption{Time-lapse snapshots of the lemniscate trajectory-tracking experiment}
  \label{fig:lemniscate trajectory tracking experiment}

\end{figure}

Time-lapse processing of the recorded experiment was conducted to visualize both the robot configuration evolution and the actual end-effector trajectory, as shown in Fig.\ref{fig:lemniscate trajectory tracking experiment}.

\section{CONCLUSIONS}

This paper presented a robust two-stage configuration planning framework that decouples the high-dimensional mobile manipulation problem into a tractable 2D base optimization. By integrating IRM with a Dijkstra-based graph search and L-BFGS refinement, kinematically feasible and smooth trajectories were successfully generated. While the current implementation operates offline, the framework achieves sub-millimeter precision in simulations and maintains a mean tracking error of 21--23 mm in physical experiments. Despite hardware-induced deviations from the Mecanum-wheeled platform, the results confirm the framework's effectiveness and practical applicability. Future work will focus on accelerating the optimization process to enable online re-planning in dynamic environments.


\end{document}